\documentclass[sn-apa]{sn-jnl}

\usepackage{svg}%
\usepackage{adjustbox}%
\usepackage{multirow}%
\usepackage{pgfplots}%
\pgfplotsset{compat=1.18}%
\usepackage{subcaption}%
\usepackage{threeparttable}%

\usepackage{graphicx}%
\usepackage{multirow}%
\usepackage{amsmath,amssymb,amsfonts}%
\usepackage{amsthm}%
\usepackage{mathrsfs}%
\usepackage[title]{appendix}%
\usepackage{xcolor}%
\usepackage{textcomp}%
\usepackage{manyfoot}%
\usepackage{booktabs}%
\usepackage{algorithm}%
\usepackage{algorithmicx}%
\usepackage{algpseudocode}%
\usepackage{listings}%

\begin{document}

\title[Evaluation of End-to-End Continuous Spanish Lipreading in Different Data Conditions]{Evaluation of End-to-End Continuous Spanish Lipreading in Different Data Conditions}

\author*[1]{\fnm{David} \sur{Gimeno-Gómez}}\email{dagigo1@dsic.upv.es}
\equalcont{}

\author[1]{\fnm{Carlos-D.} \sur{Martínez-Hinarejos}}\email{cmartine@dsic.upv.es}
\equalcont{These authors contributed equally to this work. \emph{This version of the article has been accepted for publication, after peer review (when applicable) but is not the Version of Record and does not reflect post-acceptance improvements, or any corrections. The Version of Record is available online at \href{https://doi.org/10.1007/s10579-025-09809-4}{10.1007/s10579-025-09809-4}}}

\affil[1]{\orgdiv{Pattern Recognition and Human Language Technologies Research Center}, \orgname{Universitat Politècnica de València}, \orgaddress{\street{Camino de Vera, s/n}, \city{València}, \postcode{46022}, \state{Comunitat Valenciana}, \country{Spain}}}

\abstract{Visual speech recognition remains an open research problem where different challenges must be considered by dispensing with the auditory sense, such as visual ambiguities, the inter-personal variability among speakers, and the complex modeling of silence. Nonetheless, recent remarkable results have been achieved in the field thanks to the availability of large-scale databases and the use of powerful attention mechanisms. Besides, multiple languages apart from English are nowadays a focus of interest. This paper presents noticeable advances in automatic continuous lipreading for Spanish. First, an end-to-end system based on the hybrid CTC/Attention architecture is presented. Experiments are conducted on two corpora of disparate nature, reaching state-of-the-art results that significantly improve the best performance obtained to date for both databases. In addition, a thorough ablation study is carried out, where it is studied how the different components that form the architecture influence the quality of speech recognition. Then, a rigorous error analysis is carried out to investigate the different factors that could affect the learning of the automatic system. Finally, a new Spanish lipreading benchmark is consolidated. Code and trained models are available at \url{https://github.com/david-gimeno/evaluating-end2end-spanish-lipreading}.}

\keywords{Visual Speech Recognition, Lipreading, Benchmarking, Error Analysis}

\maketitle

\section{Introduction}

In their origins, speech technologies were focused solely on acoustic cues \citep{gales2008application}. Subsequently, numerous studies demonstrating the relevance of visual cues throughout the speech perception process \citep{mcgurk1976hearing,besle2004bimodal,campbell2008processing} inspired the design of audio-visual approaches \citep{potamianos2003recent,maja2021conformers}. Nonetheless, Visual Speech Recognition (VSR), which aims to interpret speech by reading the speaker's lips, has been a focus of interest for the research community in the last few decades \citep{fernandez2018survey}. Indeed, lipreading technologies lay the groundwork for a diverse range of applications. These include silent speech passwords \citep{silent2020passwd}, which can mitigate the security vulnerabilities of more conventional text- or voice-based authentication methods, and visual keyword spotting \citep{stafylakis2018zero} for information retrieval from archival silent films \citep{jha2019spotting} or for use in forensic and criminal investigations \citep{theobald2006law,bowden2013recent}. Additionally, lipreading supports the development of assistive devices for speech-impaired individuals, particularly those unable to produce voice \citep{laux2023care,musalia2023liopa}. Other significant applications explore the integration of visual speech cues in multiple and diverse domains, including noise-robust speech recognition \citep{anwar23muavic}, sign language recognition \citep{koller2015continuous,ivanko2019signlip}, active speaker detection \citep{tao2021talknetasd,liao2023lightasd}, and lip-synced talking face synthesis for automated dubbing \citep{prajwal2020lip}. Collectively, these advancements drive progress in human-computer interaction and contribute to the development of more advanced spoken dialog systems \citep{park2024facetoface}.

VSR systems reflect a similar evolution to that observed in the field of Acoustic Speech Recognition (ASR) \citep{fernandez2018survey}, starting from traditional paradigms \citep{gales2008application,potamianos2003recent} to recent end-to-end architectures based on powerful attention mechanisms \citep{baevski2020wav2vec,maja2021conformers}. Nowadays, unprecedented high-quality recognition rates have been reached in VSR, especially in the English language \citep{prajwal2021sub,ma2022visual,shi2022learning,chang2024conformervsr}. One of the factors driving this progress has been the availability of large-scale audio-visual databases, such as LRS3-TED \citep{afouras2018lrs3}, CMU-MOSEAS \citep{zadeh2020moseas}, and MuAViC \citep{anwar23muavic}. Equally important are advances in architecture design for speech processing, including hybrid CTC/Attention decoders \citep{watanabe2017ctcattention}, self-supervised audio-visual encoders \citep{shi2022learning,haliassos2024braven}, and Transformer variants capable of processing local-context dependencies \citep{gulati20_interspeech}.

However, achieving robust performance in lipreading remains an open research problem, particularly given the unique challenges involved, such as the complex modeling of silence \citep{thangthai2018computer}, visual ambiguities \citep{bear2014phoneme,fernandez2017optimizing}, the inter-personal variability among speakers \citep{cox2008challenge}, differing light conditions, or more technical aspects, such as frame rate or image resolution \citep{bear2016decoding,bear2014resolution,dungan2018impact}. These challenges can be even more pronounced for underrepresented languages, where limited resources and development further hinders the potential of VSR applications.

While new architectures have primarily been tested in English, their application to other languages, such as Spanish, has been quite limited. Only a few studies \citep{ma2022visual,kim2023lip,yeo2024limited} have explored advanced models for this language, with most focused on TED talks \citep{afouras2018lrs3,anwar23muavic}, which, despite being in the wild, represent a relatively narrow domain. These studies are not investigating how such systems perform across diverse scenarios and varying data conditions, such as indoor studio recordings, personal vlogs, and TV newscasts. Consequently, despite the recent efforts to cover multiple languages, including Spanish, there are no well-established benchmarks that promote advances in this regard, and thus support the robust development of systems for real-world applications across different scenarios.

\noindent\textbf{Contributions.} These were the main reasons that motivated our research, whose key contributions are as follows: 

\begin{itemize}
    \item The development of end-to-end VSR systems explicitly trained for the Spanish language based on the state-of-the-art CTC/Attention architecture.

    \item The proposal of a Spanish lipreading benchmark which promotes advances in this area by covering a wide range of scenarios. Unlike prior works, this benchmark is not limited to a specific data domain, but includes diverse recording settings, speaker-independent and speaker-dependent partitions, as well as data-scarcity situations.

    \item A rigorous error analysis and an ablation study on how each architecture component, as well as the language model, affects the quality of speech recognition when dispensing with the auditory sense. This investigation aims to identify the factors that could be affecting and limiting the learning of these automatic lipreading systems.     
\end{itemize}

\section{Related Work}
\label{sec:related}

This section presents a brief description focused on the current state of the art in VSR, as well as an overview of how the Spanish language has been addressed in the field.

\noindent\textbf{Current State of the Art.} \cite{shi2022learning} introduced AV-HuBERT, a cross-modal encoder trained in a self-supervised manner using both the acoustic and visual cues. Then, once robust visual speech representations were obtained, an end-to-end VSR system was estimated after assembling a Transformer-based decoder. \cite{prajwal2021sub} not only defined an attention module especially aimed at extracting representative visual features, but also explored a subword-level recognition, arguing that it might be useful to better model visual ambiguities. \cite{ma2022visual} showed that, apart from the importance of designing an appropriate architecture through the use of Conformer encoders \citep{gulati20_interspeech} and hybrid CTC/Attention decoders \citep{watanabe2017ctcattention}, incorporating auxiliary tasks, like using enriched acoustic representations to guide the visual feature encoding, might lead to further advances in the field. Details about this architecture can be found in Section \ref{sec:vsr}. In general terms, all these works reached performances around 25-30\% WER for the English corpora LRS2-BBC \citep{son2017lrs2}, and LRS3-TED \citep{afouras2018lrs3}. Notably, various studies \citep{ma2023auto,liu2023synthvsr} have recently significantly surpassed this performance by designing methods that rely not only on models comprising vast amounts of parameters, but also on additional large-scale datasets, including synthetic video data, for their pre-training. Therefore, the current performances around 15-20\% WER are not directly comparable to our case study, which focuses on conditions with limited resources.

\noindent\textbf{Spanish Visual Speech Recognition.} \cite{fernandez2017towards} presented the VLRF corpus, whose primary purpose was assessing the VSR task's feasibility. In further research \citep{adriana2022alr}, the authors designed an end-to-end architecture, reporting results of around 72\% WER. In one of our prior works \citep{gimeno2024continuous}, we proposed a method to improve the performance of traditional HMM-based systems for VSR, achieving performances of approximately 60\% WER. Additionally, we presented the challenging LIP-RTVE database \citep{lrec2022liprtve} and proposed a traditional approach as a baseline. However, while the speaker-dependent provided around 80\% WER, acceptable results were not reached for the speaker-independent scenario, with roughly 95\% WER. Recently, multiple languages, including Spanish, were considered in \citep{ma2022visual}, achieving 56.6\% and 44.6\% WER for the Spanish partition of the MuAViC \citep{salesky21_interspeech,anwar23muavic}, and the CMU-MOSEAS \citep{zadeh2020moseas} corpora, respectively. Similarly to English, more recent pre-trained, large-scales models \citep{yeo2024limited} have been explored, surpassing the state of the art in MuAVic with results around 46\% WER. Details on all these Spanish databases considered in our proposed lipreading benchmark are described in Section \ref{sec:databases}.

\section{Model Architecture}
\label{sec:vsr}

This section describes the entire VSR system defined in our research work, covering the data preprocessing to extract our regions of interest, and the model architecture, as well as details regarding its training and inference processes.

\subsection{Data Preprocessing}

\begin{figure}[htbp]
  \centering
  \includegraphics[width=0.8\textwidth]{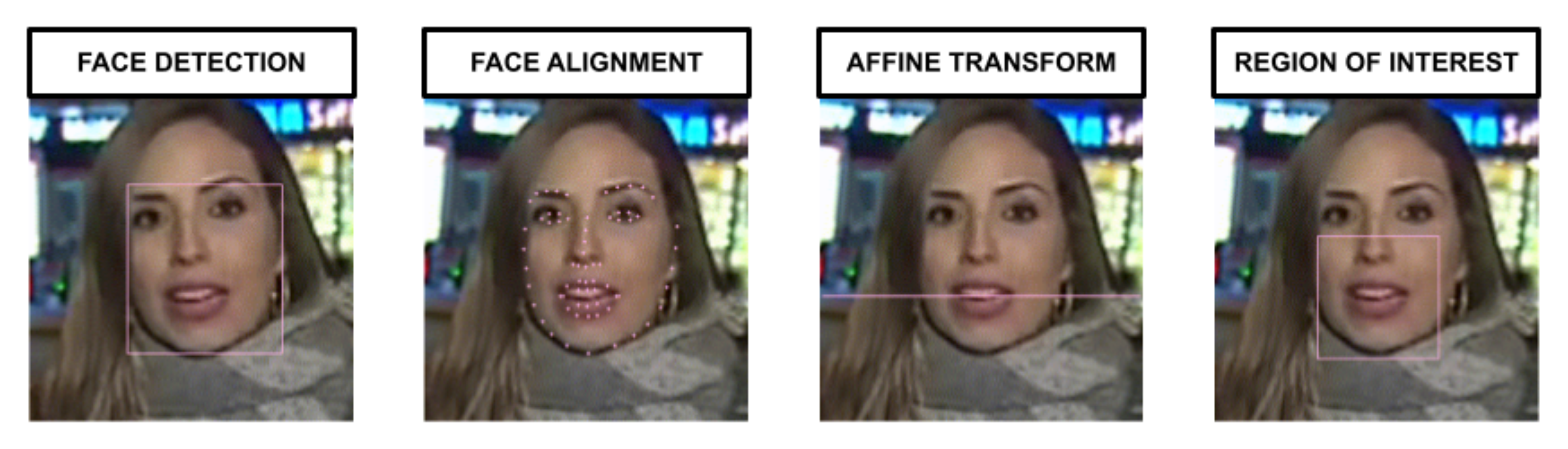}
  \caption{Data preprocessing for VSR tasks involves identifying the speaker's face, detecting 68 facial landmarks, applying an affine transformation w.r.t. a neutral reference frame to remove translation and scaling variations, and finally extracting the region of interest centered on the speaker's mouth.}
  \label{fig:preprocessing}
\end{figure}

\noindent\textbf{Video Data.} By using the RetinaFace face detector \citep{deng2020retina} and the Face Alignment Network \citep{bulat2017facealign}, we were able to extract a bounding box of 96$\times$96 pixels centered on the speaker's mouth. Subsequently, we applied a similarity transformation w.r.t. a neutral reference frame, thereby removing possible translation and scaling variations. The cropped patches are then converted to gray-scale, and normalized based on the overall mean and variance of the corresponding training set. Regarding data augmentation, random cropping of 88$\times$88 pixels, horizontal flipping, and time masking \citep{ma2022visual} are applied during the training process. These regions of interest cover the mouth, as well as the complete jaw and cheeks of the speaker, a wider area that has shown benefits when addressing the VSR task \citep{zhang2020rois}. Figure \ref{fig:preprocessing} illustrates the overall outline of our visual data processing pipeline.

\noindent\textbf{Text Data.} We normalized the transcriptions of each speech utterance through punctuation removal and lowercasing. Similar to \citep{ma2022visual}, we then used a character-level tokenizer with a vocabulary size of 37 symbols, including special ones such as the ‘space’, ’end of sentence’, and the ‘blank’ tokens.

\subsection{Visual Speech Recognition System}
\begin{figure}[htbp]
  \centering
  \includegraphics[width=\textwidth]{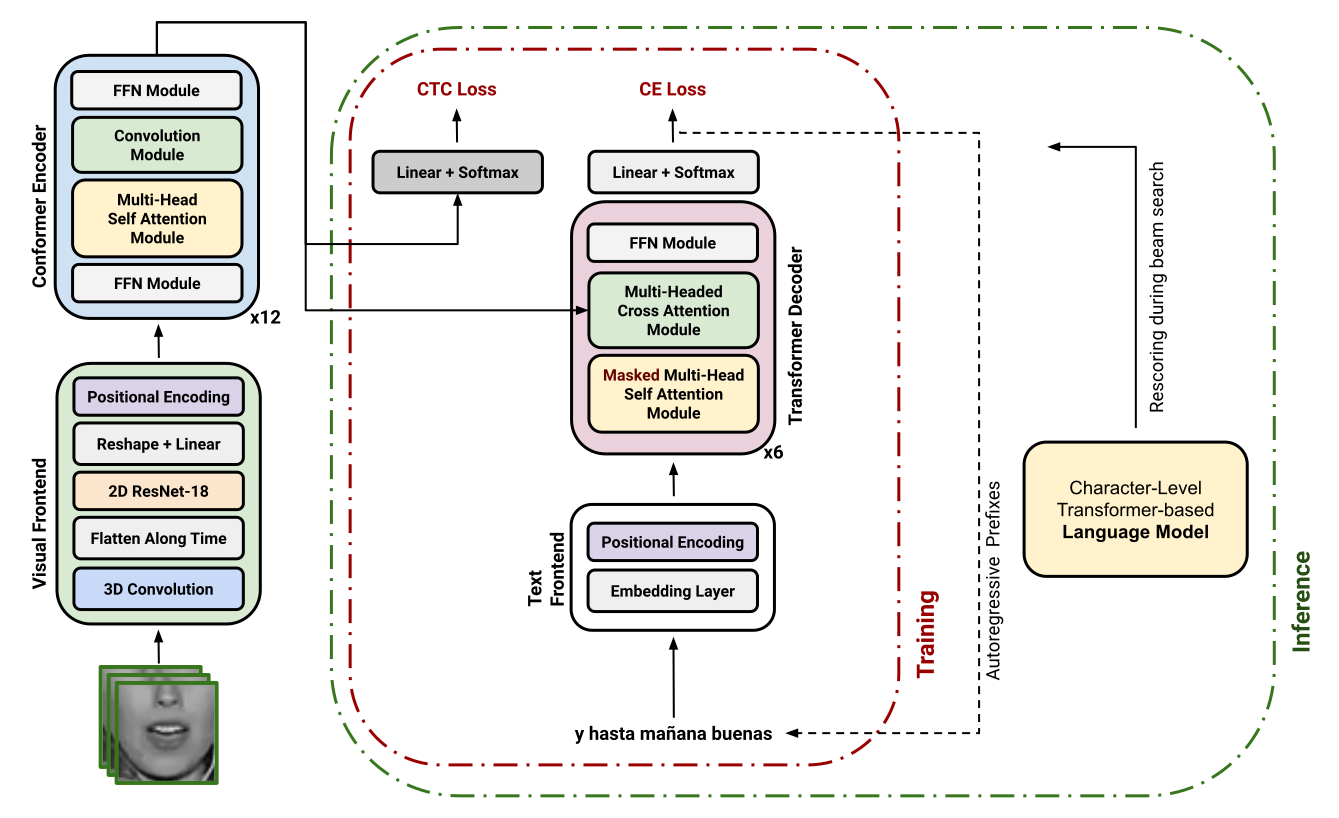}
  \caption{Architecture of the end-to-end VSR model based on the auto-regressive CTC/Attention paradigm, including design details both for its training and inference processes. For simplicity, the initial layer normalization, the residual connection, and the final dropout of each module that compose each layer of the Conformer- and Transformer-based modules are omitted. FFN, CE, and CTC refer to Feed-Forward Network, Cross Entropy, and Connectionist Temporal Classification, respectively.}
  \label{fig:e2e-scheme}
\end{figure}

The VSR system used in our work is based on the so-called CTC/Attention architecture \citep{watanabe2017ctcattention}, which represents the current state-of-the-art in the field \citep{ma2022visual}. As reflected in Figure \ref{fig:e2e-scheme}, different modules are distinguished:

\begin{itemize}
    \item \textbf{Visual Frontend:} it consists of a 2D ResNet-18 \citep{he2016resnet} whose first layer was replaced by a 3D convolutional layer in charge of dealing with temporal relationships. Specifically, this convolutional layer was defined as a kernel of size 7$\times$7 pixels with a receptive field of 5 frames. The video sequence is then reshaped to recover its temporal dimension. Subsequently, the resulting visual latent embeddings are projected into a 256-dimensional space and further injected with a relative positional encoding \citep{dai2019transformerxl}. In all layers, the Swish activation function \citep{swish2017prajit} was used. The entire visual frontend comprises about 11 million parameters.
    
    \item \textbf{Conformer Encoder:} a 12-layer Conformer encoder \citep{gulati20_interspeech} is defined to capture global and local speech interactions from the previous visual latent representation. Each layer is structured with four modules: two feed-forward networks in a macaron style, a multi-head self-attention module, and a convolution module. Layer normalization precedes each module, while a residual connection and a final dropout are applied over its output. The main difference w.r.t. the original Transformer encoder architecture \citep{vaswani2017attention} is the incorporation of a convolutional module, which is composed of point- and depth-wise convolutions. The entire temporal encoder comprises about 32 million parameters.
\end{itemize}

\begin{itemize}

    \item \textbf{Hybrid CTC/Attention Decoder:} this auto-regressive decoding paradigm is composed of a 6-layer Transformer decoder \citep{vaswani2017attention} and a fully connected layer as the CTC-based decoding branch \citep{graves2006ctc}. Combining both paradigms enables the model to adopt the Markov assumptions of CTC, aligning well with the nature of speech, while also benefiting from the flexibility of the non-sequential alignments offered by the attention-based decoder. In this architecture, the CTC-based branch directly processes the visual latent representation from the encoder, while the Transformer decoder, following its original implementation \citep{vaswani2017attention}, conditions this representation with the text data through a masked cross-attention mechanism. The entire hybrid CTC/Attention decoder comprises about 9 million parameters.

    \item{\textbf{Language Model:}} a character-level Language Model (LM) was defined as a 16-layer Transformer encoder \citep{vaswani2017attention} with a latent representation of 512 dimensions. This model comprises about 50 million parameters.
    
\end{itemize}

\subsection{Training}

The hybrid CTC/Attention architecture combines the complementary properties offered by both the CTC and the Attention paradigm, an approach that has led to advances in speech processing \citep{watanabe2017ctcattention,maja2021conformers,ma2023auto}. Accordingly, the loss function used for model training is computed as follows:

\begin{equation} \label{eq:training}
    \mathcal{L} = \alpha \log p_{ctc}(\textbf{y}|\textbf{x}) + (1 - \alpha) \log p_{attn}(\textbf{y}|\textbf{x})
\end{equation}

\noindent where $p_{ctc}$ and $p_{attn}$ denote the CTC and the Attention posteriors, respectively. In both terms, $\textbf{x}$ and $\textbf{y}$ refer to the input visual stream and its corresponding character-level target transcription, respectively. Additionally, the $\alpha$ weight is introduced to balance the relative influence of each decoder.

\subsection{Inference}

During inference, the VSR system and the Transformer-based LM were integrated in a shallow fusion manner through a beam search process, as reflected in:

\begin{equation} \label{eq:decoding}
    S = \lambda S_{ctc} + (1 - \lambda)S_{attn} + \beta S_{lm}
\end{equation}

\noindent where $S_{ctc}$ and $S_{attn}$ are the scores of the CTC and the Attention decoder, respectively. $\lambda$ is their corresponding relative weight, and $\beta$ and $S_{lm}$ refer to the LM decoding influence weight and the LM score, respectively. In this inference process, it is important to note that the Attention-based decoder primarily determines when the decoding concludes by predicting the end-of-sentence token, and thus the CTC-based decoding branch serves as an additional scoring factor \citep{watanabe2017ctcattention}.

\section{Databases}
\label{sec:databases}

To promote research on Spanish VSR, in addition to including the Spanish corpora explored in \citep{ma2022visual}, we propose two additional databases of different natures to consolidate a new Spanish lipreading benchmark that covers a wide range of data conditions. Table \ref{tab:databases} highlights the main differences between these corpora and emphasizes how the model architecture proposed in this work is capable of adapting to each different scenario, as discussed in Section \ref{sec:results}.

\begin{figure}[!htbp]
  \centering
  \includegraphics[width=0.55\textwidth]{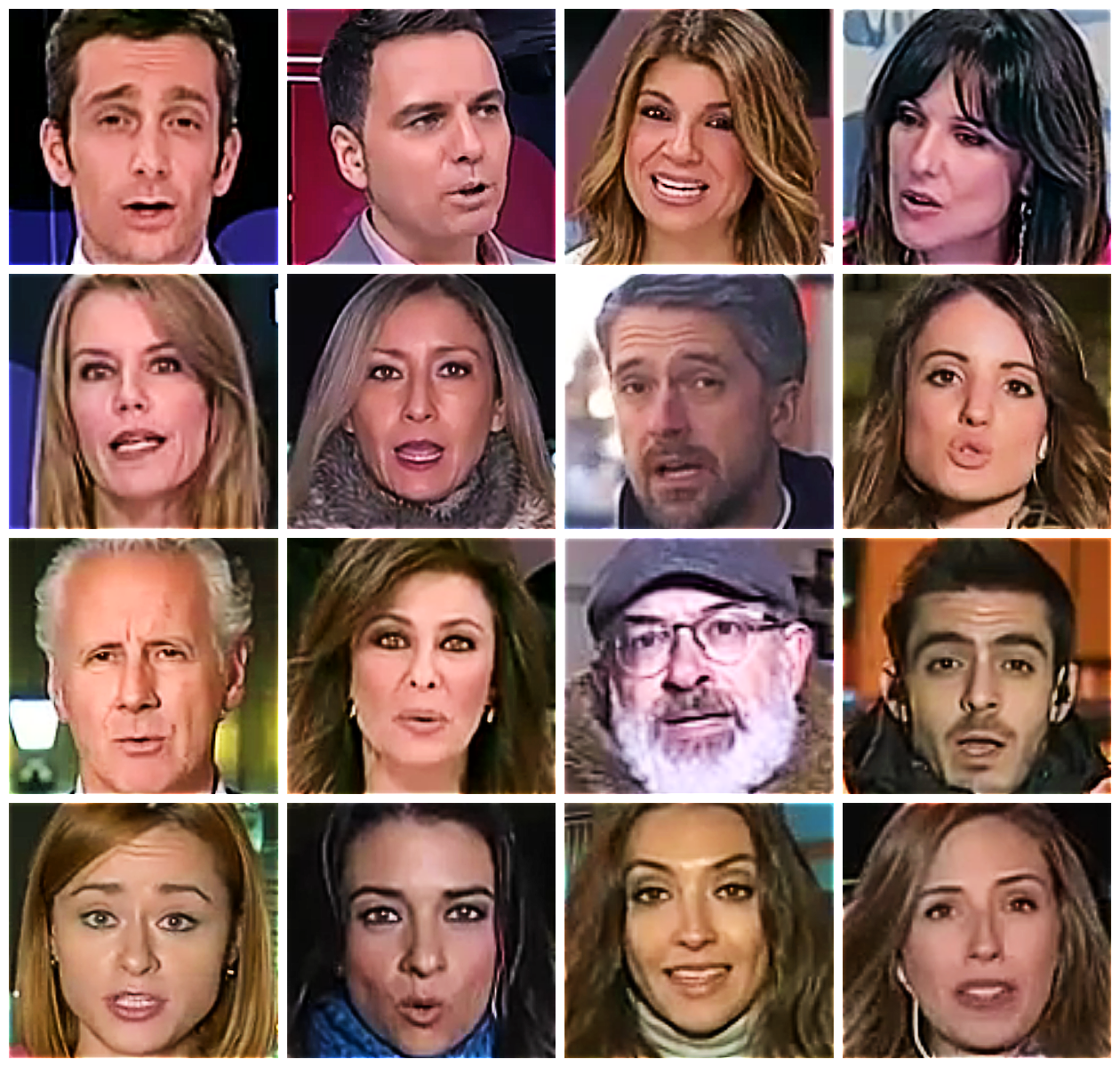}
  \caption{Excerpts from original videos in the LIP-RTVE database, showcasing the diverse speakers and scenarios to be addressed in this challenging task.}
  \label{fig:liprtve-showcasing}
\end{figure}

\begin{itemize}

    \item \textbf{VLRF} \citep{fernandez2017towards} is a database that aims to study the feasibility of VSR. Therefore, although it addresses continuous speech, it should be noted that the corpus was recorded in controlled settings, and speakers were asked to strive to be understood. This speaker-dependent database offers a training set with 480 samples (1 hour) and a test set with 120 samples (14 minutes). It provides around 4k running words with a vocabulary size of 1373 different words.
    
    \item \textbf{LIP-RTVE} \citep{lrec2022liprtve} is a challenging database collected from TV newscast programs, thus offering proper support to estimate VSR systems for realistic scenarios, as Figure \ref{fig:liprtve-showcasing} reflects. In addition, partitions for both speaker-dependent (SD) and speaker-independent (SI) scenarios are provided. The SI partition defines a training set with 7,142 samples (9 hours), a validation set with 1638 samples (2 hours), and a test set with 1572 samples (2 hours). The SD partition offers a training set with 7,355 samples (9 hours), a validation set with 1,597 samples (2 hours), and a test set with 1400 samples (2 hours). The database provides more than 140k running words with a vocabulary size of around 10k different words.
\end{itemize}

\begin{itemize}

    \item \textbf{CMU-MOSEAS\textsubscript{es}} \citep{zadeh2020moseas} is the Spanish subset of an audiovisual database collected from YouTube monologue videos covering multiple languages, namely Portuguese, French, Spanish, and German. In our work, we only considered its Spanish partition. According to \cite{ma2022visual}, we defined a training set with 8,253 samples (15.7 hours) and a test set with 329 samples (0.6 hours). The database provides more than 150k running words with a vocabulary size of around 14k different words.

    \item \textbf{MuAViC\textsubscript{es}} is the Spanish language partition of a large-scale multilingual corpus collected from TED talks. It is based on and follows the data splits defined by the Multilingual-TEDx dataset \citep{salesky21_interspeech}\footnote{Although previous works, such as \citep{ma2022visual}, refer to Multilingual-TEDx, as MuAViC had not been publicly released at that time, we focused on MuAVic since it is now more popular and well-established.}. The Spanish partition consists of a training set with 102,035 samples (177.5 hours), a validation set with 906 (1.6 hours), and a test set with 1012 samples (1.7 hours). The database provides over 1.7M running words with a vocabulary size of around 58k different words.

\end{itemize}

\begin{table}[!htbp]
\footnotesize
\setlength\tabcolsep{3.5pt}
\caption{\label{tab:databases}
Comparison of the proposed audio-visual Spanish databases}
\centering
\begin{tabular}{lcccc}
\toprule
\textbf{} & \textbf{VLRF} & \textbf{LIP-RTVE} & \textbf{CMU-MOSEAS\textsubscript{es}} & \textbf{MuAViC\textsubscript{es}}\\
\toprule
\textbf{Duration} & $\sim$1 hour & $\sim$13 hours & $\sim$16 hours & $\sim$181 hours \\
\textbf{No. Speakers} & 24  & 323 & 341 & 1016 \\
\textbf{Vocabulary} & 1,373 words & 9,308 words & 14,447 words & 57,692 words\\
\textbf{Nature} & Controlled Recording & TV Newscasts & YouTube Vlogs & TEDx Talks \\
\textbf{Frame Rate} & 50 fps & 25 fps & 25 fps & 25 fps \\
\textbf{Resolution} & 1280$\times$720 pixels & 480$\times$270 pixels & 1280$\times$720 pixels & 1280$\times$720 pixels \\
\bottomrule
\end{tabular}
\end{table}

The rationale behind the proposed lipreading benchmark for Spanish is to evaluate VSR systems across a wide range of scenarios, addressing the heterogeneity and occasional significant lack of data that most languages suffer from in the context of audiovisual speech technologies. According to our prior research work \citep{acosta2024annotheia}, although there are numerous audio-visual datasets\footnote{A more comprehensive review of existing audiovisual databases can be found in \citep{fernandez2018survey} and \citep{acosta2024annotheia}} for word-level classification tasks \citep{chung2017lip,lrw2019chinese,egorov2021lrwr} or those collected in controlled recording studios \citep{harte2015tcd}, only a few languages have support for continuous, in-the-wild audiovisual speech recognition. The recent MuAViC \citep{anwar23muavic} is one the first multi-lingual datasets in this regard. However, it presents a significant inequality in terms of the number of hours among the different languages. Therefore, our work aims to promote robust benchmarks across diverse scenarios to ensure that lipreading technologies can be applied in multiple data conditions to which future researchers and developers might be limited.

\section{Experimental Setup}
\label{sec:setup}

\subsection{Implementation Details}
\label{sec:implementation}

The implementation of our VSR models is based on the ESPNet toolkit \citep{watanabe18_interspeech}. Experiments were conducted on a 12-core 3.50GHz Intel i7-7800X CPU and a GeForce RTX 2080 GPU with 8GB memory.

\noindent\textbf{Pre-Training Stage.} Both the VSR system and the Transformer-based LM were initially pre-trained using the weights publicly released by \cite{ma2022visual} for the Spanish language. The VSR system was pre-trained with more than 1500 hours of data from different English corpora to subsequently be fine-tuned using the Spanish partition of the Multilingual-TEDx \citep{salesky21_interspeech} and the CMU-MOSEAS \citep{zadeh2020moseas} databases, achieving recognition rates of around 56.6\% and 44.6\% WER, respectively. The LM can be considered a general Spanish LM, since it has been trained over a total of 192 million characters collected from the Spanish Multilingual TEDx \citep{salesky21_interspeech}, Common Voice \citep{ardila2020common}, and Multilingual LibriSpeech \citep{pratap20interspeech} corpora.

\noindent\textbf{Fine-Tuning Data Sets.} When fine-tuning the VSR systems, the official splits described in Section \ref{sec:databases} were used both for the VLRF and LIP-RTVE databases. We also studied fine-tuning the LMs. Therefore, the nearly 300k sentences collected by \cite{adriana2022alr} were used for the VLRF corpus. In the case of the LIP-RTVE database, around 80k sentences collected from different but contemporary TV newscasts \citep{lrec2022liprtve} were considered.

\noindent\textbf{VSR Training Setup.} The AdamW optimizer \citep{loshchilov2017decoupled} and a linear one-cycle learning rate scheduler \citep{leslie2019onecycle} were used during 5 epochs. For the VLRF, the learning rate was set to 5$\times$10\textsuperscript{-4}. For the LIP-RTVE, a value of 5$\times$10\textsuperscript{-4} and 5$\times$10\textsuperscript{-5} were set to the SD and SI scenarios, respectively. In all cases, a batch size of 1 sample had to be established due to our GPU memory limitations (see Subsection \ref{sec:experiments}). Regarding the CTC/Attention loss function, the $\alpha$ weight specified in Equation (\ref{eq:training}) was set to 0.1, according to the settings specified in \citep{ma2022visual}.

\noindent\textbf{VSR Inference Setup.} As reflected in Equation (\ref{eq:decoding}), different weights model the influence of each component during inference. In those experiments where the LM was not discarded, the $\beta$ weight was set to 0.4, as specified in \citep{ma2022visual}. Conversely, when it was discarded, it meant setting $\beta$ to 0.0. Regarding the CTC/Attention balance, the $\lambda$ weight was set to 0.1. Besides, the word insertion penalty was set to 0.0, while the beam size was set to 10. 

\noindent\textbf{LM Fine-Tuning.} LMs were fine-tuned using the optimizer and scheduler described for the VSR system, except for setting a learning rate of 5$\times$10\textsuperscript{-5} over 5 epochs.

\noindent\textbf{Evaluation Metric.} Results reported in our work were evaluated using the well-known Word Error Rate (WER) with 95\% confidence intervals obtained by the bootstrap method described by \cite{bisani2004bootstrap}.

\subsection{Experiments}
\label{sec:experiments}

In this paper, we introduce new state-of-the-art results for the VLRF and LIP-RTVE datasets. While our primary contributions focus on these datasets, we also incorporate results from other corpora to establish a comprehensive Spanish lipreading benchmark, aiming to drive broader development in Spanish VSR technologies.

\noindent\textbf{Ablation Study.} We explored the impact and contribution of various modules and learning techniques that make up the entire model architecture over the overall system performance. To do so, our study consisted of independently removing one of these model components, including the data augmentation, language model, as well as both the CTC and Attention-based decoding branches. It should be noted that when one of the decoder branches, either the CTC- or Attention-based, was discarded, it was discarded during both training and inference. Therefore, if the CTC decoder was removed, the parameters $\alpha$ (Equation (\ref{eq:training})) and $\lambda$ (Equation (\ref{eq:decoding})) were both set to 0.0. Conversely, if the Attention decoder was discarded, $\alpha$ and $\lambda$ were set to 1.0. Similarly, if the LM was discarded, the parameter $\beta$ in Equation (\ref{eq:decoding}) was set to 0.0. A discussion about the results of this ablation study can be found in Section \ref{sec:results}.

\noindent\textbf{Limitations.} The main limitations have been related to the requirement for large GPU resources. Firstly, we were not able to study how the incorporation of auxiliary tasks, as \cite{ma2022visual} did in their research, could influence the learning of the VSR system. Besides, it was not possible to explore larger batch sizes which might have led to a more stable estimation and faster convergence. It should be noted that we explored the accumulating gradient strategy \citep{ott2018accum}. However, no significant differences were found. We argue that despite applying this technique, the normalization layers were still affected by the actual reduced batch size.

\section{Results \& Discussion}
\label{sec:results}

\noindent\textbf{Ablation Study on VLRF.} As reflected in Table \ref{tab:ablation}, the proposed approach achieves, in the best setting along our experiments, around 25\% WER. This is a noticeable result that has significantly outperformed the previous state of the art in the VLRF task, which reached a 59.7\% WER \citep{gimeno2024continuous}.

Regarding the ablation study, it was reasonable that skipping the fine-tuning process would lead to unacceptable results since the nature of the VLRF corpus differs considerably from the databases employed in \cite{ma2022visual}. Besides, the VLRF corpus does not belong to a specific domain, but it was compiled from a subset of pre-defined sentences. This could explain why fine-tuning the LM with more general-Spanish text did not yield substantial differences.

\noindent\textbf{Ablation Study on LIP-RTVE.} Compared to the results discussed in Section \ref{sec:related}, remarkable performances on the LIP-RTVE database, which could be considered as the new state of the art in the task, have been achieved. Specifically, our proposed approach has reached, in its best setting along our experiments, absolute WER improvements of around 45\% and 35\% for the SD and the SI scenario, respectively. 

Regarding the ablation study, we should take into account that the LIP-RTVE database presents, unlike the VLRF corpus, a more similar nature to the data used in \citep{ma2022visual}. Consequently, the results obtained by skipping the fine-tuning process were, although bad, more acceptable. In addition, as might be expected, the SD performance is more affected when this process is not performed. In addition, since the LIP-RTVE belongs to the specific domain of TV newscast programmes, we observed how fine-tuning the LM significantly improved the system performance. Notably, the LM presents a stronger influence in the SD scenario, whether fine-tuned or not. The different complexity between both scenarios might be the cause. During the inference phase, the VSR system proposes several alternative transcriptions, each with its corresponding score computed. If these alternatives are of insufficient quality, as might be the case with the SI setting, the score provided by the LM cannot contribute significantly, as none of the alternatives stands out over the other.

\begin{table}[!htbp]
\footnotesize
\setlength\tabcolsep{2.4pt}
\caption{Ablation study on the proposed Spanish databases. Details on Subsection \ref{sec:experiments}. LM refers to Language Model, CTC to Connectionist Temporal Classification, and $\Delta$ to the absolute error increment w.r.t the entire model architecture.}
\centering
\begin{tabular}{@{}lccccccccc@{}}
\toprule
\multicolumn{1}{c}{} &  & \multicolumn{2}{c}{\textbf{\begin{tabular}[c]{@{}c@{}}VLRF\\ speaker-dependent\end{tabular}}} &  & \multicolumn{2}{c}{\textbf{\begin{tabular}[c]{@{}c@{}}LIP-RTVE\\ speaker-dependent\end{tabular}}} &  & \multicolumn{2}{c}{\textbf{\begin{tabular}[c]{@{}c@{}}LIP-RTVE\\ speaker-independent\end{tabular}}}\\
\cmidrule{3-4} \cmidrule{6-7} \cmidrule{9-10}
\multicolumn{1}{c}{} &  & \textbf{\%WER} & \textbf{$\Delta$} &  & \textbf{\%WER} & \textbf{$\Delta$}  &  & \textbf{\%WER} & \textbf{$\Delta$} \\ \midrule
\multicolumn{1}{c}{\textbf{Entire Model Architecture}} &  & 24.8$\pm$3.4 & - &  & 34.5$\pm$1.2 & - &  & 59.5$\pm$1.2 & - \\
\qquad$-$ Data Augmentation &  & 23.9$\pm$3.4 & $-$0.9 &  & 32.3$\pm$1.2 & $-$2.2 &  & 58.7$\pm$1.2 & $-$0.8 \\
\qquad$-$ LM Fine-Tuning &  & 26.2$\pm$3.2 & $+$1.4 &  & 38.7$\pm$1.2 & $+$4.2 &  & 61.9$\pm$1.1 & $+$2.4 \\
\qquad$-$ Language Model &  & 35.9$\pm$3.0 & $+$11.1 &  & 41.4$\pm$1.2 & $+$6.9 &  & 64.8$\pm$1.0 & $+$5.3 \\
\qquad$-$ Attention Decoder &  & 41.7$\pm$3.9 & $+$16.9 &  & 53.1$\pm$1.3 & $+$18.6 &  & 68.6$\pm$1.1 & $+$9.1 \\
\qquad$-$ CTC Decoder &  & 51.3$\pm$3.9 & $+$26.5 &  & 70.1$\pm$1.1 & $+$35.6 &  & 75.4$\pm$1.0 & $+$15.9 \\
\qquad$-$ Fine-Tuning &  & $>$100\tnote{$\dagger$} & $+\infty$ &  & 78.3$\pm$1.2 & $+$43.8 &  & 74.4$\pm$1.2 & $+$14.1 \\ \bottomrule
\end{tabular}
\begin{tablenotes}
    \footnotesize
    \item[$\dagger$] due to a peculiarity of the WER metric, results above 100\% were obtained 
\end{tablenotes}
\label{tab:ablation}
\end{table}

\noindent\textbf{Overall Analysis.} By studying all the experiments carried out in our research from a broader point of view, we could observe, although with different proportions, a common behaviour or pattern. Hence, we were able to conclude that:

\begin{itemize}
    \item Our experiments support the hypothesis that the explored VSR system architecture was capable of adapting to various data conditions across multiple domains, scenarios and data availability settings.
    
    \item The employed data augmentation process seems to be inadequate for our data since it did not provide significant differences in terms of performance for both the VLRF and the LIP-RTVE. We could state that the use of this technique slightly hinders the learning of the system when fine-tuning for a limited number of epochs.

    \item In general terms, the LM presents a significant influence over the quality of speech recognition. Nonetheless, depending on certain aspects that we previously mentioned, fine-tuning the LM does not always offer remarkable improvements. 
    
    \item Regarding the hybrid CTC/Attention decoder, CTC stands as one of the fundamental factors of our VSR system. When this decoder branch is discarded, the system performance suffers a drastic deterioration in all our experiments. Conversely, if the attention decoder is removed, i.e., an architecture based solely on the CTC paradigm, a lower deterioration can be observed. This finding may seem counterintuitive, as the lack of visible speech information is addressed through context modeling \citep{fernandez2018survey}, a property highly exploited by attention-based mechanisms \citep{vaswani2017attention}. Nonetheless, our experimental results support the work carried out in \citep{watanabe2017ctcattention} in the sense that Markov assumptions provided by the CTC paradigm \citep{graves2006ctc} significantly contribute to the overall system quality. Furthermore, our results are consistent with \citep{ma2022visual}, supporting that the combination of the CTC paradigm with Attention-based decoders provides a multiobjective learning framework for addressing convergence misalignment issues in the context of automatic lipreading.    
\end{itemize}

\begin{figure}[!htbp]
\centering
\captionsetup{font=small}
\begin{adjustbox}{max width=0.9\textwidth}

\begin{subfigure}[a]{8cm}
\begin{tikzpicture}
\begin{axis}[height=5cm,
             width=8cm,
             xlabel=\%WER,
             xmin=-5, xmax=105,
             xtick distance=10,
             ylabel=Frequency,
             ymin=0, ymax=40,
             minor y tick num = 4,
             area style]
\addplot+[ybar interval, mark=no, draw=black!60!green, fill=green!30] coordinates { (0, 35) (10, 32) (20, 20) (30, 9) (40, 10) (50, 4) (60, 5) (70, 2) (80, 1) (90, 1) (100, 0)};
\end{axis}
\end{tikzpicture}
\caption{\large The speaker-dependent VLRF} \label{fig:vlrf-hist}
\end{subfigure}

\begin{subfigure}[b]{8cm}
\begin{tikzpicture}
\begin{axis}[height=5cm,
             width=8cm,
             xlabel=\%WER,
             xmin=-5, xmax=105,
             xtick distance=10,
             ylabel=Frequency,
             ymin=0, ymax=400,
             minor y tick num = 4,
             area style]

\addplot+[ybar interval, mark=no, draw=black!60!blue, fill=blue!30] coordinates { (0, 370) (10, 189) (20, 190) (30, 172) (40, 149) (50, 96) (60, 72) (70, 52) (80, 29) (90, 81) (100, 0)};
\end{axis}
\end{tikzpicture}
\caption{\large The speaker-dependent LIP-RTVE} \label{fig:liprtvesd-hist}
\end{subfigure}

\begin{subfigure}[c]{8cm}
\begin{tikzpicture}
\begin{axis}[height=5cm,
             width=8cm,
             xlabel=\%WER,
             xmin=-5, xmax=105,
             xtick distance=10,
             ylabel=Frequency,
             ymin=0, ymax=300,
             minor y tick num = 4,
             area style]
\addplot+[ybar interval, mark=no, draw=black!60!red, fill=red!30] coordinates { (0, 103) (10, 75) (20, 117) (30, 189) (40, 210) (50, 160) (60, 152) (70, 174) (80, 117) (90, 275) (100, 0)};
\end{axis}
\end{tikzpicture}
\caption{\large The speaker-independent LIP-RTVE} \label{fig:liprtvesi-hist}
\end{subfigure}
\end{adjustbox}
\caption{System performance (\%WER) histogram of the proposed Spanish databases.} \label{fig:hists}

\end{figure}
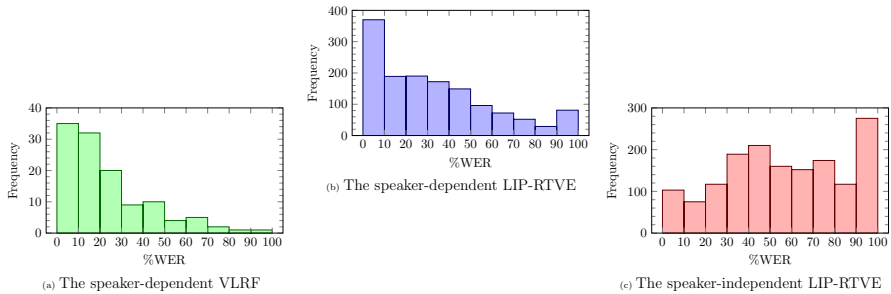

\noindent\textbf{Error Analysis.} In order to better understand the reasons behind the performance of our VSR systems, different error analyses were conducted. Table \ref{tab:examples} shows examples of how our VSR systems interpret speech. Multiple mistakes are found in most cases, but some of them seem to be related to visual ambiguities \citep{bear2014phoneme,fernandez2017optimizing}. Figure \ref{fig:hists} reflects histograms w.r.t. the number of samples included in a certain WER range for each database partition. We also studied how the word length affected the system performance. However, we were not able to identify any trend or pattern in this regard.

\begin{table}[!htbp]
\centering
\setlength\tabcolsep{5.0pt}
\footnotesize
\caption{Examples of the proposed Spanish VSR systems and their corresponding performances in terms of Word Error rate (WER) and Character Error Rate (CER). SD and SI refer to the speaker-dependent and speaker-independent partition of the corresponding database, respectively. \emph{ref} and \emph{hyp} indicate the reference (ground truth) and the hypothesis provided by the automatic lipreading system, respectively.}
\begin{tabular}{@{}crlcc@{}}
\toprule
\textbf{Database} & & \textbf{Transcription} & \textbf{\%WER} & \textbf{\%CER} \\ \midrule

\multirow{9}{*}{\textbf{\shortstack{VLRF \\ (SD)}}} & \textbf{ref:} & el chino vino a la escuela de intercambio & \multirow{3}{*}{50.0} & \multirow{3}{*}{19.5} \\
& & (\emph{the chinese came to the exchange school}) & &  \\
 & \textbf{hyp:} & sino vino a la escuela de este cambio & &  \\\cmidrule(lr){2-5}
 & \textbf{ref:} & tu hermano y el mio se encontraron en el metro & \multirow{3}{*}{20.0} & \multirow{3}{*}{8.7} \\
 & & (\emph{your brother and mine met on the subway}) & &  \\
 & \textbf{hyp:} & tu hermano y el vino se encontraron en el medio & & \\\cmidrule(lr){2-5}
 & \textbf{ref:} & la pelicula que vimos era una comedia & \multirow{3}{*}{0.0} & \multirow{3}{*}{0.0} \\
 & & (\emph{the film we saw was a comedy}) & &  \\
 & \textbf{hyp:} & la pelicula que vimos era una comedia & & \\\midrule\midrule
 
\multirow{9}{*}{\textbf{\shortstack{LIP-RTVE \\ (SD)}}} & \textbf{ref:} & pena con hasta tres años de prision & \multirow{3}{*}{71.4} & \multirow{3}{*}{28.6} \\
& & (\emph{penalty of up to three years in prison}) & &  \\
 & \textbf{hyp:} & pero esta traslados de prision &  &  \\\cmidrule(lr){2-5}
 & \textbf{ref:} & y hasta mañana muy buenas noches & \multirow{3}{*}{33.3} & \multirow{3}{*}{12.5} \\
 & & (\emph{and until tomorrow very good night}) & &  \\
 & \textbf{hyp:} & esta mañana muy buenas noches &  & \\\cmidrule(lr){2-5}
 & \textbf{ref:} & a partir de mañana lunes a las doce de la noche & \multirow{3}{*}{0.0} & \multirow{3}{*}{0.0} \\
 & & (\emph{starting tomorrow Monday at midnight}) & &  \\
 & \textbf{hyp:} & a partir de mañana lunes a las doce de la noche &  & \\ \midrule\midrule
 
 \multirow{9}{*}{\textbf{\shortstack{LIP-RTVE \\ (SI)}}} & \textbf{ref:} & se le aparece en la cabeza una imagen & \multirow{2}{*}{75.0} & \multirow{3}{*}{45.9} \\
 & & (\emph{an image appears in your head}) & &  \\
 & \textbf{hyp:} & aparece que dice una imagen &  &  \\\cmidrule(lr){2-5}
 & \textbf{ref:} & estan limpiando tambien el barro y evaluando los destrozos & \multirow{3}{*}{66.7} & \multirow{3}{*}{32.8} \\
 & & (\emph{they are also cleaning the mud and evaluating the damage}) & &  \\
 & \textbf{hyp:} & se esta inspirando tambien el perro y evaluando seis socios &  & \\\cmidrule(lr){2-5}
 & \textbf{ref:} & cumplen un mes en prision & \multirow{3}{*}{20.0} & \multirow{3}{*}{4.0} \\
 & & (\emph{they spend a month in prison}) & &  \\
 & \textbf{hyp:} & cumple un mes en prision & & \\ \bottomrule
\end{tabular}
\label{tab:examples}
\end{table}

In addition, as it has been an object of study for decades in the field of statistical linguistics \citep{piantadosi2014zipf}, we decided to consider Zipf's law \citep{zipf1936law,zipf1949human} in our error analysis. According to this law, the word frequencies are inversely proportional to their frequency rank, i.e., the most frequent word presents a frequency proportional to $1$, the second most frequent word presents a frequency proportional to $1/2$, the third most frequent word presents a frequency proportional to $1/3$, and so forth. This behaviour practically holds across all human languages \citep{piantadosi2014zipf}, including those artificially constructed, such as the case of Esperanto \citep{manaris2006esperanto}. Besides, this law is a valuable reference for language learning and foreign language teaching, since a certain percentage of the most frequent words would represent most of the content of the language \citep{feng2023formal}.

\pgfplotstableread[col sep=comma]{zipf_train_vlrf_sd.csv}\vlrf
\pgfplotstableread[col sep=comma]{zipf_train_liprtve_sd.csv}\liprtvesd
\pgfplotstableread[col sep=comma]{zipf_train_liprtve_si.csv}\liprtvesi
\begin{figure}[!htbp]
\centering
\begin{tikzpicture}
\begin{axis}[width=0.6\textwidth,
             ymode=log,
             log basis y=2,
             xmode=log,
             log basis x=2,
             legend cell align={left},
             legend style={at={(0.03,0.03)}, anchor={south west}, font=\scriptsize},
             ylabel=Relative Frequency,
             xlabel=Frequency Rank,
             ]
\addplot[mark=no, dashed, black!60, ultra thick,] table [x=vocab_id, y=zipf_law, col sep=comma] {\liprtvesd}; \addlegendentry{Zipf's law}

\addplot[mark=no, mark options={solid}, black!10!red, ultra thick,] table [x=vocab_id, y=database_rank, col sep=comma] {\vlrf}; \addlegendentry{VLRF (SD)}

\addplot[mark=no, mark options={solid}, black!10!blue, ultra thick,] table [x=vocab_id, y=database_rank, col sep=comma] {\liprtvesd}; \addlegendentry{LIP-RTVE (SD)}

\addplot[mark=no, mark options={solid}, black!10!yellow, ultra thick,] table [x=vocab_id, y=database_rank, col sep=comma] {\liprtvesi}; \addlegendentry{LIP-RTVE (SI)}

\end{axis}
\end{tikzpicture}
\caption{Relationship between Zipf's law and the VLRF and LIP-RTVE databases. Similar behaviour was observed for the Spanish CMU-MOSEAS and MuAViC corpora, but they were omitted for clarity. SD and SI refer to a speaker-dependent and speaker-independent partition of the corresponding database, respectively. It should be noted that both dimensions are depicted in a logarithmic scale.}
\label{fig:zipf}
\end{figure}
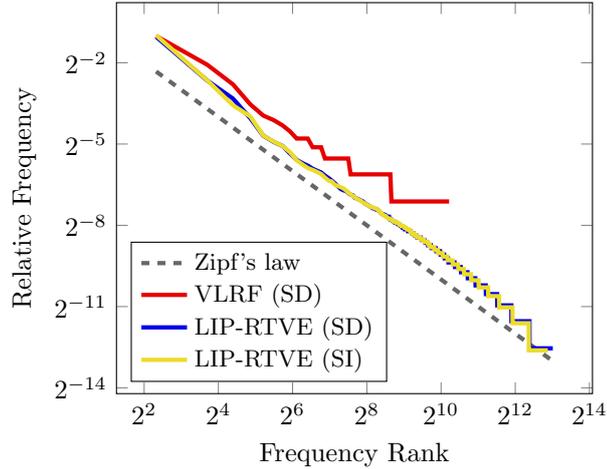
\let\vlrf\relax
\let\liprtvesd\relax
\let\liprtvesi\relax

Therefore, because automatic speech recognition is meant to interpret language, we considered it interesting to analyze how the proposed Spanish databases for this paper relate to Zipf's law. Hence, for each database and using its corresponding training set, we computed the relative frequency of each vocabulary word w.r.t. the most frequent word. Figure \ref{fig:zipf} shows that, although the databases studied in our work are conditioned by the purpose why they were collected or the domain to which they belong, all of them approximately hold Zipf's law. For the VLRF, the first 500 words with the highest frequency cover 77.1\% of the total running words of training. For the rest of the proposed databases, the 1000 most frequent words cover a similar 77.8\% of the total running words of training. However, we can observe that the word frequency is higher than expected w.r.t. the ideal Zipf's law, which could be a consequence of the scarce data offered by the proposed databases or the consequence of having addressed specific domains where the vocabulary could be limited and condensed. Table \ref{tab:zipf} reflects the conducted further analyses to study how different aspects related to Zipf's law could have been affecting our VSR systems performance:

\begin{itemize}
     \item When comparing both partitions of the LIP-RTVE database, we can observe how the different statistics reflect why the SD scenario is an easier task. Although the percentage of test vocabulary words covered by the top highest-frequency words from the training set (\textbf{test-v $\cap$ top-v}) is slightly lower in this scenario, there is a significant difference in the rest of the computed statistics. First, we see how the SI training set is less representative of the test set in terms of vocabulary (\textbf{test-v $\cap$ train-v}) and running words (\textbf{test-rw $\cap$ train-v}). Secondly, we also observed that the test running words are more covered by the top highest-frequency words (\textbf{test-rw $\cap$ top-v}) in the SD scenario. These aspects suggest that the SD partition would be affected to a lesser extent by the out-of-vocabulary problem.
     
\end{itemize}

\begin{table}[!htbp]
\centering
\setlength\tabcolsep{2pt}
\caption{Analysis on the relationship between the vocabulary words of the training set (\textbf{train-v}) and test set (\textbf{test-v}) of each database partition w.r.t. the top highest-frequency words (\textbf{top-v}) of the corresponding database. In the case of VLRF, \textbf{top-v} refers to the 500 highest-frequency vocabulary words, while for the rest of the proposed databases, it includes the first 1000 highest-frequency vocabulary words. \textbf{$\cap$} represents the intersection operation. Reported percentages were computed w.r.t. the first data set of the intersection. In addition, recognition rates obtained for each database are reported in terms of Word Error Rate (WER).}
\label{tab:zipf}
\begin{tabular}{@{}lccccccc@{}}
\toprule
 \multicolumn{1}{l}{} & & \multicolumn{2}{c}{\textbf{Speaker Dependent}} & & \multicolumn{3}{c}{\textbf{Speaker Independent}} \\
 \cmidrule{3-4} \cmidrule{6-8}
 & &  \textbf{VLRF} & \textbf{LIP-RTVE} & & \textbf{CMU-MOSEAS\textsubscript{es}} & \textbf{MuAViC\textsubscript{es}} & \textbf{LIP-RTVE} \\ \midrule
\textbf{train-v} & & 1195 & 8244 & & 14126 & 57129 & 7524 \\
\textbf{test-v} & & 418 & 4133 & & 1769 & 3200 & 2983 \\
\textbf{test-rw} & & 779 & 21259 & & 6064 & 15412 & 20133 \\\midrule
\textbf{test-v $\cap$ train-v} & & 240 (57.4\%) & 3574 (86.5\%) & & 1418 (80.2\%) & 2959 (92.5\%) & 2026 (67.9\%)\\
\textbf{test-v $\cap$ top-v} & & 135 (32.3\%) & 961 (23.3\%) & & 649 (36.7\%) & 823 (25.7\%) & 773 (25.9\%)\\
\textbf{test-rw $\cap$ train-v} & & 588 (75.5\%) & 20626 (97.0\%) & & 5648 (93.1\%) & 15073 (97.8\%) & 18312 (91.0\%) \\
\textbf{test-rw $\cap$ top-v} & & 473 (60.7\%) & 16483 (77.5\%) & & 4597 (75.8\%) & 11438 (74.2\%) & 15025 (74.6\%)\\\midrule\midrule
\textbf{\%WER} & & \textbf{24.8$\pm$3.4} & \textbf{34.5$\pm$1.2} & & \textbf{44.6$\pm$0.6}\tnote{$\dagger$} & \textbf{56.3$\pm$0.3}\tnote{$\dagger$} & \textbf{59.5$\pm$1.2}\\\bottomrule
\end{tabular}
\begin{tablenotes}
    \footnotesize
    \item[$\dagger$] performance reported by \cite{ma2022visual}. 
\end{tablenotes}
\end{table}

\begin{itemize}
     \item For SI dataset partitions, we realized how the differences in terms of coverage of test vocabulary and running words correlate to task complexity and system performance degradation. Comparing MuAViC and LIP-RTVE, both databases reflect similar test vocabulary coverage w.r.t. the top highest-frequency word, which may explain why their respective system performances do not differ substantially. However, we can observe a noticeable reduction of LIP-RTVE in terms of test set coverage when contrasting relative to the training set. If we then contrast these statistics with those computed for CMU-MOSEAS, we can infer that the coverage by the top highest-frequency words is one of the most determinant factors in achieving better recognition rates. As previously discussed, these factors relate to the out-of-vocabulary problem. Although slight, this difference could cause more errors to occur when decoding speech, which, in turn, could generate other errors in a cascade over the autoregressive beam search process. Therefore, we conclude that LIP-RTVE is the most challenging task in our proposed benchmark, not only because of the limited amount of training data it offers, but also because of its extensive test set comprising a broader vocabulary.

     \item Although we are seeing that the coverage of test vocabulary and running words is an important aspect related to the VSR system performance, the comparison of CMU-MOSEAS with the SD dataset partitions, or even the comparison between VLRF and the SD partition of the LIP-RTVE, suggest that other factors may also be influencing this performance. However, we cannot ignore that, despite most of the statistics cover notably lower percentages, the VLRF task achieves better recognition rates than the LIP-RTVE SD corpus. Although the VLRF's test set indeed presents 9\% more vocabulary words in common with the top highest-frequency words, this finding suggests that other aspects related to the inherent challenges of multi-speaker lipreading, such as recording in controlled scenarios, the frame rate, the image resolution or the speed of speech, could be responsible for this system performance difference we are discussing.
\end{itemize}

All these findings support the notion that VSR is a challenging task influenced by multiple factors. Additionally, our error analysis shows the relevance of Zipf's law when we train automatic visual speech recognition systems.
\newline

\noindent\textbf{Spanish Lipreading Benchmark.} A summary of both the work proposed in this paper and the experiments in Spanish conducted by \cite{ma2022visual} is presented in Table \ref{tab:benchmark} for benchmarking purposes. Furthermore, as previously discussed in Section \ref{sec:databases}, we consider databases of different natures covering diverse data conditions to ensure robust evaluation that aligns with the heterogeneity and occasional significant lack of data that most languages suffer from in the context of audiovisual speech technologies \citep{acosta2024annotheia}. Hence, our work studies how a VSR architecture is capable of adapting to different characteristics and scenarios. By consolidating this benchmark, we aimed to promote further research in the field of Spanish VSR.

Table \ref{tab:benchmark} shows the results obtained using the VSR system proposed in \citep{ma2022visual}, a model that was first pre-trained on more than 1500 hours of English data and then adapted to Spanish. However, these results are still not comparable to the English state-of-the-art, which is around 25-30\% WER. While our error analysis has shed light on this performance discrepancy, further advances in the field, as \cite{ma2022visual} advocate, might be achieved not only through the utilization of larger databases, but also by carefully designing model architectures capable of handling multiple languages within the context of continuous VSR.

\begin{table}[!htbp]
\setlength\tabcolsep{5pt}
\centering
\caption{The proposed Spanish Lipreading Benchmark.}
\begin{tabular}{@{}ccclccc@{}}
\toprule
 & \multicolumn{2}{c}{\textbf{Speaker Dependent}} &  & \multicolumn{3}{c}{\textbf{Speaker Independent}} \\
 \cmidrule{2-3} \cmidrule{5-7}
 & \textbf{VLRF} & \textbf{LIP-RTVE} &  & \textbf{CMU-MOSEAS\textsubscript{es}} & \textbf{MuAViC\textsubscript{es}} & \textbf{LIP-RTVE} \\ \midrule
\textbf{\%WER} & 24.8$\pm$3.4 & 34.5$\pm$1.2 &  & 44.6$\pm$0.6\tnote{$\dagger$} & 56.3$\pm$0.3\tnote{$\dagger$} & 59.5$\pm$1.2 \\ \bottomrule
\end{tabular}
\begin{tablenotes}
    \footnotesize
    \item[$\dagger$] performance reported by \cite{ma2022visual}. 
\end{tablenotes}
\label{tab:benchmark}
\end{table}

\vspace{-0.5cm}

\section{Conclusions \& Future Work}
\label{sec:conclusions}

In this work, noticeable advances in Spanish VSR have been achieved. We employed a model based on the hybrid CTC/Attention architecture \citep{watanabe2017ctcattention} to obtain state-of-the-art results, significantly outperforming the best results reported to date for two Spanish databases of disparate nature. Additionally, a thorough ablation study was carried out to investigate how the different components that form the architecture influence the quality of speech recognition. Specifically, apart from several aspects related to the LM or the reasons behind certain behaviours, our study placed the CTC paradigm as one of the most influential factors in our VSR system. Our error analysis further supported the relevance of Zipf's law when estimating automatic lipreading systems and how it can affect the overall model performance. Finally, we established a new Spanish lipreading benchmark to promote research in the field.

Regarding future work, one of our intentions is to explore how audio cues might be incorporated into the development of audio-visual speech recognition systems for Spanish. Furthermore, we plan to investigate non-autoregressive architectures \citep{higuchi2021maskctc,lee2021interctc}, as well as parameter-efficient techniques \citep{fernandez2023sparsevsr,wei2023sim}, thus aiming at the development of VSR systems for real-world applications.

\section*{Statements and Declarations}

\bmhead{Funding}

This work was partially supported by Grant CIACIF/2021/295 funded by Generalitat Valenciana and by Grant PID2021-124719OB-I00 under project LLEER (PID2021-124719OB-100) funded by MCIN/AEI/10.13039/501100011033/ and by ERDF, EU A way of making Europe.

\bmhead{Conflict of interest} 

The authors declare no conflict of interest. The funders had no role in the design of the study; in the collection, analyses, or interpretation of data; in the writing of the manuscript; or in the decision to publish the results.

\bmhead{Ethics approval}

Not applicable.

\bmhead{Code availability} 

Code and trained models are available at \url{https://github.com/david-gimeno/evaluating-end2end-spanish-lipreading}.

\bibliography{main}

\begin{thebibliography}{}
\renewcommand{\doi}[1]{\url{https://doi.org/#1}}
\bibcommenthead

\bibitem [\protect \citeauthoryear {%
Acosta-Triana%
, Gimeno-Gómez%
\BCBL {}\ \BBA {} Martínez-Hinarejos%
}{%
Acosta-Triana%
\ \protect \BOthers {.}}{%
{\protect \APACyear {2024}}%
}]{%
acosta2024annotheia}
\APACinsertmetastar {%
acosta2024annotheia}%
\begin{APACrefauthors}%
Acosta-Triana, J\BHBI M.%
, Gimeno-Gómez, D.%
\BCBL {} Martínez-Hinarejos, C\BHBI D.%
\end{APACrefauthors}%
\unskip\
\newblock
\APACrefYearMonthDay{2024}{}{}.
\newblock
{\BBOQ}\APACrefatitle {{AnnoTheia: A Semi-Automatic Annotation Toolkit for
  Audio-Visual Speech Technologies}} {{AnnoTheia: A Semi-Automatic Annotation
  Toolkit for Audio-Visual Speech Technologies}}.{\BBCQ}
\newblock
 \APACrefbtitle {{Proc. of LREC-COLING}} {{Proc. of LREC-COLING}}\ (\BPGS\
  1260--1269).
\PrintBackRefs{\CurrentBib}

\bibitem [\protect \citeauthoryear {%
Afouras%
, Chung%
\BCBL {}\ \BBA {} Zisserman%
}{%
Afouras%
\ \protect \BOthers {.}}{%
{\protect \APACyear {2018}}%
}]{%
afouras2018lrs3}
\APACinsertmetastar {%
afouras2018lrs3}%
\begin{APACrefauthors}%
Afouras, T.%
, Chung, J\BHBI S.%
\BCBL {} Zisserman, A.%
\end{APACrefauthors}%
\unskip\
\newblock
\APACrefYearMonthDay{2018}{}{}.
\newblock
{\BBOQ}\APACrefatitle {L{RS3}-{TED}: a large-scale dataset for visual speech
  recognition} {L{RS3}-{TED}: a large-scale dataset for visual speech
  recognition}.{\BBCQ}
\newblock
\APACjournalVolNumPages{arXiv preprint arXiv:1809.00496}{}{}{,}
\newblock
\begin{APACrefURL} {\url{https://doi.org/10.48550/arXiv.1809.00496}}
  \end{APACrefURL}
\newblock

\newblock

\PrintBackRefs{\CurrentBib}

\bibitem [\protect \citeauthoryear {%
Anwar%
\ \protect \BOthers {.}}{%
Anwar%
\ \protect \BOthers {.}}{%
{\protect \APACyear {2023}}%
}]{%
anwar23muavic}
\APACinsertmetastar {%
anwar23muavic}%
\begin{APACrefauthors}%
Anwar, M.%
, Shi, B.%
, Goswami, V.%
, Hsu, W.%
, Pino, J.%
\BCBL {} Wang, C.%
\end{APACrefauthors}%
\unskip\
\newblock
\APACrefYearMonthDay{2023}{}{}.
\newblock
{\BBOQ}\APACrefatitle {{MuAViC: A Multilingual Audio-Visual Corpus for Robust
  Speech Recognition and Robust Speech-to-Text Translation}} {{MuAViC: A
  Multilingual Audio-Visual Corpus for Robust Speech Recognition and Robust
  Speech-to-Text Translation}}.{\BBCQ}
\newblock
 \APACrefbtitle {Interspeech} {Interspeech}\ (\BPGS\ 4064--4068).
\PrintBackRefs{\CurrentBib}

\bibitem [\protect \citeauthoryear {%
Ardila%
\ \protect \BOthers {.}}{%
Ardila%
\ \protect \BOthers {.}}{%
{\protect \APACyear {2020}}%
}]{%
ardila2020common}
\APACinsertmetastar {%
ardila2020common}%
\begin{APACrefauthors}%
Ardila, R.%
, Branson, M.%
, Davis, K.%
, Kohler, M.%
, Meyer, J.%
, Henretty, M.%
\BDBL {}Weber, G.%
\end{APACrefauthors}%
\unskip\
\newblock
\APACrefYearMonthDay{2020}{}{}.
\newblock
{\BBOQ}\APACrefatitle {Common Voice: A Massively-Multilingual Speech Corpus}
  {Common voice: A massively-multilingual speech corpus}.{\BBCQ}
\newblock
 \APACrefbtitle {{Proc. LREC}} {{Proc. LREC}}\ (\BPGS\ 4218--4222).
\newblock
\begin{APACrefURL} {\url{https://aclanthology.org/2020.lrec-1.520}}
  \end{APACrefURL}
\PrintBackRefs{\CurrentBib}

\bibitem [\protect \citeauthoryear {%
Baevski%
, Zhou%
, Mohamed%
\BCBL {}\ \BBA {} Auli%
}{%
Baevski%
\ \protect \BOthers {.}}{%
{\protect \APACyear {2020}}%
}]{%
baevski2020wav2vec}
\APACinsertmetastar {%
baevski2020wav2vec}%
\begin{APACrefauthors}%
Baevski, A.%
, Zhou, Y.%
, Mohamed, A.%
\BCBL {} Auli, M.%
\end{APACrefauthors}%
\unskip\
\newblock
\APACrefYearMonthDay{2020}{}{}.
\newblock
{\BBOQ}\APACrefatitle {wav2vec 2.0: A framework for self-supervised learning of
  speech representations} {wav2vec 2.0: A framework for self-supervised
  learning of speech representations}.{\BBCQ}
\newblock
\APACjournalVolNumPages{Advances in neural information processing
  systems}{33}{}{12449--12460,}
\newblock
\begin{APACrefDOI} \doi{10.5555/3495724.3496768} \end{APACrefDOI}
\newblock

\newblock

\PrintBackRefs{\CurrentBib}

\bibitem [\protect \citeauthoryear {%
Bear%
\ \BBA {} Harvey%
}{%
Bear%
\ \BBA {} Harvey%
}{%
{\protect \APACyear {2016}}%
}]{%
bear2016decoding}
\APACinsertmetastar {%
bear2016decoding}%
\begin{APACrefauthors}%
Bear, H.%
\BCBT {}\ \BBA {} Harvey, R.%
\end{APACrefauthors}%
\unskip\
\newblock
\APACrefYearMonthDay{2016}{}{}.
\newblock
{\BBOQ}\APACrefatitle {Decoding visemes: Improving machine lip-reading}
  {Decoding visemes: Improving machine lip-reading}.{\BBCQ}
\newblock
 \APACrefbtitle {I{CASSP}} {I{CASSP}}\ (\BPGS\ 2009--2013).
\newblock
\begin{APACrefURL} {\url{https://doi.org/10.1109/ICASSP.2016.7472029}}
  \end{APACrefURL}
\PrintBackRefs{\CurrentBib}

\bibitem [\protect \citeauthoryear {%
Bear%
, Harvey%
, Theobald%
\BCBL {}\ \BBA {} Lan%
}{%
Bear%
\ \protect \BOthers {.}}{%
{\protect \APACyear {2014}}%
{\protect \APACexlab {{\protect \BCnt {1}}}}}]{%
bear2014resolution}
\APACinsertmetastar {%
bear2014resolution}%
\begin{APACrefauthors}%
Bear, H.%
, Harvey, R.%
, Theobald, B.%
\BCBL {} Lan, Y.%
\end{APACrefauthors}%
\unskip\
\newblock
\APACrefYearMonthDay{2014{\protect \BCnt {1}}}{}{}.
\newblock
{\BBOQ}\APACrefatitle {Resolution limits on visual speech recognition}
  {Resolution limits on visual speech recognition}.{\BBCQ}
\newblock
 \APACrefbtitle {I{CIP}} {I{CIP}}\ (\BPGS\ 1371--1375).
\newblock
\begin{APACrefURL} {\url{https://doi.org/10.1109/ICIP.2014.7025274}}
  \end{APACrefURL}
\PrintBackRefs{\CurrentBib}

\bibitem [\protect \citeauthoryear {%
Bear%
, Harvey%
, Theobald%
\BCBL {}\ \BBA {} Lan%
}{%
Bear%
\ \protect \BOthers {.}}{%
{\protect \APACyear {2014}}%
{\protect \APACexlab {{\protect \BCnt {2}}}}}]{%
bear2014phoneme}
\APACinsertmetastar {%
bear2014phoneme}%
\begin{APACrefauthors}%
Bear, H.%
, Harvey, R.%
, Theobald, B.%
\BCBL {} Lan, Y.%
\end{APACrefauthors}%
\unskip\
\newblock
\APACrefYearMonthDay{2014{\protect \BCnt {2}}}{}{}.
\newblock
{\BBOQ}\APACrefatitle {Which phoneme-to-viseme maps best improve visual-only
  computer lip-reading?} {Which phoneme-to-viseme maps best improve visual-only
  computer lip-reading?}{\BBCQ}
\newblock
 \APACrefbtitle {International Symposium on Visual Computing} {International
  symposium on visual computing}\ (\BPGS\ 230--239).
\newblock
\begin{APACrefURL} {\url{https://doi.org/10.1007/978-3-319-14364-4_22}}
  \end{APACrefURL}
\PrintBackRefs{\CurrentBib}

\bibitem [\protect \citeauthoryear {%
Besle%
, Fort%
, Delpuech%
\BCBL {}\ \BBA {} Giard%
}{%
Besle%
\ \protect \BOthers {.}}{%
{\protect \APACyear {2004}}%
}]{%
besle2004bimodal}
\APACinsertmetastar {%
besle2004bimodal}%
\begin{APACrefauthors}%
Besle, J.%
, Fort, A.%
, Delpuech, C.%
\BCBL {} Giard, M\BHBI H.%
\end{APACrefauthors}%
\unskip\
\newblock
\APACrefYearMonthDay{2004}{}{}.
\newblock
{\BBOQ}\APACrefatitle {Bimodal speech: early suppressive visual effects in
  human auditory cortex} {Bimodal speech: early suppressive visual effects in
  human auditory cortex}.{\BBCQ}
\newblock
\APACjournalVolNumPages{European journal of Neuroscience}{20}{8}{2225--2234,}
\newblock
\begin{APACrefDOI} \doi{10.1111\%2Fj.1460-9568.2004.03670.x} \end{APACrefDOI}
\newblock

\newblock

\PrintBackRefs{\CurrentBib}

\bibitem [\protect \citeauthoryear {%
Bisani%
\ \BBA {} Ney%
}{%
Bisani%
\ \BBA {} Ney%
}{%
{\protect \APACyear {2004}}%
}]{%
bisani2004bootstrap}
\APACinsertmetastar {%
bisani2004bootstrap}%
\begin{APACrefauthors}%
Bisani, M.%
\BCBT {}\ \BBA {} Ney, H.%
\end{APACrefauthors}%
\unskip\
\newblock
\APACrefYearMonthDay{2004}{}{}.
\newblock
{\BBOQ}\APACrefatitle {Bootstrap estimates for confidence intervals in ASR
  performance evaluation} {Bootstrap estimates for confidence intervals in asr
  performance evaluation}.{\BBCQ}
\newblock
 \APACrefbtitle {ICASSP} {Icassp}\ (\BVOL~1, \BPGS\ 409--412).
\newblock
\begin{APACrefURL} {\url{https://doi.org/10.1109/ICASSP.2004.1326009}}
  \end{APACrefURL}
\PrintBackRefs{\CurrentBib}

\bibitem [\protect \citeauthoryear {%
Bowden%
\ \protect \BOthers {.}}{%
Bowden%
\ \protect \BOthers {.}}{%
{\protect \APACyear {2013}}%
}]{%
bowden2013recent}
\APACinsertmetastar {%
bowden2013recent}%
\begin{APACrefauthors}%
Bowden, R.%
, Cox, S.%
, Harvey, R.%
, Lan, Y.%
, Ong, E\BHBI J.%
, Owen, G.%
\BCBL {} Theobald, B\BHBI J.%
\end{APACrefauthors}%
\unskip\
\newblock
\APACrefYearMonthDay{2013}{}{}.
\newblock
{\BBOQ}\APACrefatitle {Recent Developments in Automated Lip-Reading} {Recent
  developments in automated lip-reading}.{\BBCQ}
\newblock
\APACjournalVolNumPages{Optics and Photonics for Counterterrorism, Crime
  Fighting and Defence IX; and Optical Materials and Biomaterials in Security
  and Defence Systems Technology X}{8901}{}{179--191,}
\newblock

\newblock

\PrintBackRefs{\CurrentBib}

\bibitem [\protect \citeauthoryear {%
Bulat%
\ \BBA {} Tzimiropoulos%
}{%
Bulat%
\ \BBA {} Tzimiropoulos%
}{%
{\protect \APACyear {2017}}%
}]{%
bulat2017facealign}
\APACinsertmetastar {%
bulat2017facealign}%
\begin{APACrefauthors}%
Bulat, A.%
\BCBT {}\ \BBA {} Tzimiropoulos, G.%
\end{APACrefauthors}%
\unskip\
\newblock
\APACrefYearMonthDay{2017}{}{}.
\newblock
{\BBOQ}\APACrefatitle {How Far are We from Solving the 2D \& 3D Face Alignment
  Problem? (and a Dataset of 230,000 3D Facial Landmarks)} {How far are we from
  solving the 2d \& 3d face alignment problem? (and a dataset of 230,000 3d
  facial landmarks)}.{\BBCQ}
\newblock
 \APACrefbtitle {ICCV} {Iccv}\ (\BPG~1021-1030).
\newblock
\begin{APACrefURL} {\url{https://doi.org/10.1109/ICCV.2017.116}}
  \end{APACrefURL}
\PrintBackRefs{\CurrentBib}

\bibitem [\protect \citeauthoryear {%
Campbell%
}{%
Campbell%
}{%
{\protect \APACyear {2008}}%
}]{%
campbell2008processing}
\APACinsertmetastar {%
campbell2008processing}%
\begin{APACrefauthors}%
Campbell, R.%
\end{APACrefauthors}%
\unskip\
\newblock
\APACrefYearMonthDay{2008}{}{}.
\newblock
{\BBOQ}\APACrefatitle {{The Processing of Audio-Visual Speech: Empirical and
  Neural Bases}} {{The Processing of Audio-Visual Speech: Empirical and Neural
  Bases}}.{\BBCQ}
\newblock
\APACjournalVolNumPages{Philosophical Transactions of the Royal Society B:
  Biological Sciences}{363}{1493}{1001--1010,}
\newblock

\newblock

\PrintBackRefs{\CurrentBib}

\bibitem [\protect \citeauthoryear {%
Chang%
, Liao%
, Serdyuk%
, Shah%
\BCBL {}\ \BBA {} Siohan%
}{%
Chang%
\ \protect \BOthers {.}}{%
{\protect \APACyear {2024}}%
}]{%
chang2024conformervsr}
\APACinsertmetastar {%
chang2024conformervsr}%
\begin{APACrefauthors}%
Chang, O.%
, Liao, H.%
, Serdyuk, D.%
, Shah, A.%
\BCBL {} Siohan, O.%
\end{APACrefauthors}%
\unskip\
\newblock
\APACrefYearMonthDay{2024}{}{}.
\newblock
{\BBOQ}\APACrefatitle {{Conformer is All You Need for Visual Speech
  Recognition}} {{Conformer is All You Need for Visual Speech
  Recognition}}.{\BBCQ}
\newblock
 \APACrefbtitle {ICASSP} {Icassp}\ (\BPG~10136-10140).
\PrintBackRefs{\CurrentBib}

\bibitem [\protect \citeauthoryear {%
Chung%
\ \BBA {} Zisserman%
}{%
Chung%
\ \BBA {} Zisserman%
}{%
{\protect \APACyear {2017}}%
}]{%
chung2017lip}
\APACinsertmetastar {%
chung2017lip}%
\begin{APACrefauthors}%
Chung, J.%
\BCBT {}\ \BBA {} Zisserman, A.%
\end{APACrefauthors}%
\unskip\
\newblock
\APACrefYearMonthDay{2017}{}{}.
\newblock
{\BBOQ}\APACrefatitle {Lip reading in the wild} {Lip reading in the
  wild}.{\BBCQ}
\newblock
 \APACrefbtitle {13th Asian Conference on Computer Vision} {13th asian
  conference on computer vision}\ (\BPGS\ 87--103).
\PrintBackRefs{\CurrentBib}

\bibitem [\protect \citeauthoryear {%
Cox%
, Harvey%
, Lan%
, Newman%
\BCBL {}\ \BBA {} Theobald%
}{%
Cox%
\ \protect \BOthers {.}}{%
{\protect \APACyear {2008}}%
}]{%
cox2008challenge}
\APACinsertmetastar {%
cox2008challenge}%
\begin{APACrefauthors}%
Cox, S.J.%
, Harvey, R.W.%
, Lan, Y.%
, Newman, J.L.%
\BCBL {} Theobald, B\BHBI J.%
\end{APACrefauthors}%
\unskip\
\newblock
\APACrefYearMonthDay{2008}{}{}.
\newblock
{\BBOQ}\APACrefatitle {The challenge of multispeaker lip-reading.} {The
  challenge of multispeaker lip-reading.}{\BBCQ}
\newblock
 \APACrefbtitle {A{VSP}} {A{VSP}}\ (\BPGS\ 179--184).
\newblock
\begin{APACrefURL}
  {\url{https://www.isca-speech.org/archive_open/avsp08/av08_179.html}}
  \end{APACrefURL}
\PrintBackRefs{\CurrentBib}

\bibitem [\protect \citeauthoryear {%
Dai%
\ \protect \BOthers {.}}{%
Dai%
\ \protect \BOthers {.}}{%
{\protect \APACyear {2019}}%
}]{%
dai2019transformerxl}
\APACinsertmetastar {%
dai2019transformerxl}%
\begin{APACrefauthors}%
Dai, Z.%
, Yang, Z.%
, Yang, Y.%
, Carbonell, J.%
, Le, Q.%
\BCBL {} Salakhutdinov, R.%
\end{APACrefauthors}%
\unskip\
\newblock
\APACrefYearMonthDay{2019}{}{}.
\newblock
{\BBOQ}\APACrefatitle {Transformer-{XL}: Attentive Language Models beyond a
  Fixed-Length Context} {Transformer-{XL}: Attentive language models beyond a
  fixed-length context}.{\BBCQ}
\newblock
 \APACrefbtitle {Proc. of the 57th ACL} {Proc. of the 57th acl}\ (\BPGS\
  2978--2988).
\newblock
\APACaddressPublisher{}{ACL}.
\PrintBackRefs{\CurrentBib}

\bibitem [\protect \citeauthoryear {%
Deng%
, Guo%
, Ververas%
, Kotsia%
\BCBL {}\ \BBA {} Zafeiriou%
}{%
Deng%
\ \protect \BOthers {.}}{%
{\protect \APACyear {2020}}%
}]{%
deng2020retina}
\APACinsertmetastar {%
deng2020retina}%
\begin{APACrefauthors}%
Deng, J.%
, Guo, J.%
, Ververas, E.%
, Kotsia, I.%
\BCBL {} Zafeiriou, S.%
\end{APACrefauthors}%
\unskip\
\newblock
\APACrefYearMonthDay{2020}{}{}.
\newblock
{\BBOQ}\APACrefatitle {RetinaFace: Single-Shot Multi-Level Face Localisation in
  the Wild} {Retinaface: Single-shot multi-level face localisation in the
  wild}.{\BBCQ}
\newblock
 \APACrefbtitle {{CVPR}} {{CVPR}}\ (\BPG~5202-5211).
\newblock
\begin{APACrefURL} {\url{10.1109/CVPR42600.2020.00525}} \end{APACrefURL}
\PrintBackRefs{\CurrentBib}

\bibitem [\protect \citeauthoryear {%
Dungan%
, Karaali%
\BCBL {}\ \BBA {} Harte%
}{%
Dungan%
\ \protect \BOthers {.}}{%
{\protect \APACyear {2018}}%
}]{%
dungan2018impact}
\APACinsertmetastar {%
dungan2018impact}%
\begin{APACrefauthors}%
Dungan, L.%
, Karaali, A.%
\BCBL {} Harte, N.%
\end{APACrefauthors}%
\unskip\
\newblock
\APACrefYearMonthDay{2018}{}{}.
\newblock
{\BBOQ}\APACrefatitle {The Impact of Reduced Video Quality on Visual Speech
  Recognition} {The impact of reduced video quality on visual speech
  recognition}.{\BBCQ}
\newblock
 \APACrefbtitle {I{CIP}} {I{CIP}}\ (\BPG~2560-2564).
\newblock
\begin{APACrefURL} {\url{http://doi.org/10.1109/ICIP.2018.8451754}}
  \end{APACrefURL}
\PrintBackRefs{\CurrentBib}

\bibitem [\protect \citeauthoryear {%
Egorov%
, Kostyumov%
, Konyk%
\BCBL {}\ \BBA {} Kolesnikov%
}{%
Egorov%
\ \protect \BOthers {.}}{%
{\protect \APACyear {2021}}%
}]{%
egorov2021lrwr}
\APACinsertmetastar {%
egorov2021lrwr}%
\begin{APACrefauthors}%
Egorov, E.%
, Kostyumov, V.%
, Konyk, M.%
\BCBL {} Kolesnikov, S.%
\end{APACrefauthors}%
\unskip\
\newblock
\APACrefYearMonthDay{2021}{}{}.
\newblock
{\BBOQ}\APACrefatitle {{LRWR: large-scale benchmark for lip reading in Russian
  language}} {{LRWR: large-scale benchmark for lip reading in Russian
  language}}.{\BBCQ}
\newblock
\APACjournalVolNumPages{arXiv preprint arXiv:2109.06692}{}{}{,}
\newblock

\newblock

\PrintBackRefs{\CurrentBib}

\bibitem [\protect \citeauthoryear {%
Ezz%
, Mostafa%
\BCBL {}\ \BBA {} Nasr%
}{%
Ezz%
\ \protect \BOthers {.}}{%
{\protect \APACyear {2020}}%
}]{%
silent2020passwd}
\APACinsertmetastar {%
silent2020passwd}%
\begin{APACrefauthors}%
Ezz, M.%
, Mostafa, A.M.%
\BCBL {} Nasr, A.A.%
\end{APACrefauthors}%
\unskip\
\newblock
\APACrefYearMonthDay{2020}{}{}.
\newblock
{\BBOQ}\APACrefatitle {{A Silent Password Recognition Framework Based on Lip
  Analysis}} {{A Silent Password Recognition Framework Based on Lip
  Analysis}}.{\BBCQ}
\newblock
\APACjournalVolNumPages{IEEE Access}{8}{}{55354-55371,}
\newblock

\newblock

\PrintBackRefs{\CurrentBib}

\bibitem [\protect \citeauthoryear {%
Feng%
}{%
Feng%
}{%
{\protect \APACyear {2023}}%
}]{%
feng2023formal}
\APACinsertmetastar {%
feng2023formal}%
\begin{APACrefauthors}%
Feng, Z.%
\end{APACrefauthors}%
\unskip\
\newblock
\APACrefYear{2023}.
\newblock
\APACrefbtitle {Formal Analysis for Natural Language Processing: A Handbook}
  {Formal analysis for natural language processing: A handbook}.
\newblock
\APACaddressPublisher{}{Springer Nature}.
\newblock
\begin{APACrefURL} {\url{https://doi.org/10.1007/978-981-16-5172-4}}
  \end{APACrefURL}
\PrintBackRefs{\CurrentBib}

\bibitem [\protect \citeauthoryear {%
Fernandez-Lopez%
\ \protect \BOthers {.}}{%
Fernandez-Lopez%
\ \protect \BOthers {.}}{%
{\protect \APACyear {2023}}%
}]{%
fernandez2023sparsevsr}
\APACinsertmetastar {%
fernandez2023sparsevsr}%
\begin{APACrefauthors}%
Fernandez-Lopez, A.%
, Chen, H.%
, Ma, P.%
, Haliassos, A.%
, Petridis, S.%
\BCBL {} Pantic, M.%
\end{APACrefauthors}%
\unskip\
\newblock
\APACrefYearMonthDay{2023}{}{}.
\newblock
{\BBOQ}\APACrefatitle {{Sparsevsr: Lightweight and noise robust visual speech
  recognition}} {{Sparsevsr: Lightweight and noise robust visual speech
  recognition}}.{\BBCQ}
\newblock
\APACjournalVolNumPages{{arXiv preprint arXiv:2307.04552}}{}{}{,}
\newblock

\newblock

\PrintBackRefs{\CurrentBib}

\bibitem [\protect \citeauthoryear {%
Fernandez-Lopez%
, Martinez%
\BCBL {}\ \BBA {} Sukno%
}{%
Fernandez-Lopez%
\ \protect \BOthers {.}}{%
{\protect \APACyear {2017}}%
}]{%
fernandez2017towards}
\APACinsertmetastar {%
fernandez2017towards}%
\begin{APACrefauthors}%
Fernandez-Lopez, A.%
, Martinez, O.%
\BCBL {} Sukno, F.M.%
\end{APACrefauthors}%
\unskip\
\newblock
\APACrefYearMonthDay{2017}{}{}.
\newblock
{\BBOQ}\APACrefatitle {Towards estimating the upper bound of visual-speech
  recognition: The visual lip-reading feasibility database} {Towards estimating
  the upper bound of visual-speech recognition: The visual lip-reading
  feasibility database}.{\BBCQ}
\newblock
 \APACrefbtitle {12th FG} {12th fg}\ (\BPGS\ 208--215).
\newblock
\begin{APACrefURL} {\url{https://doi.org/10.1109/FG.2017.34}} \end{APACrefURL}
\PrintBackRefs{\CurrentBib}

\bibitem [\protect \citeauthoryear {%
Fernandez-Lopez%
\ \BBA {} Sukno%
}{%
Fernandez-Lopez%
\ \BBA {} Sukno%
}{%
{\protect \APACyear {2018}}%
}]{%
fernandez2018survey}
\APACinsertmetastar {%
fernandez2018survey}%
\begin{APACrefauthors}%
Fernandez-Lopez, A.%
\BCBT {}\ \BBA {} Sukno, F.M.%
\end{APACrefauthors}%
\unskip\
\newblock
\APACrefYearMonthDay{2018}{}{}.
\newblock
{\BBOQ}\APACrefatitle {Survey on automatic lip-reading in the era of deep
  learning} {Survey on automatic lip-reading in the era of deep
  learning}.{\BBCQ}
\newblock
\APACjournalVolNumPages{Image and Vision Computing}{78}{}{53--72,}
\newblock
\begin{APACrefDOI} \doi{https://doi.org/10.1016/j.imavis.2018.07.002}
  \end{APACrefDOI}
\newblock

\newblock

\PrintBackRefs{\CurrentBib}

\bibitem [\protect \citeauthoryear {%
Fernandez-Lopez%
\ \BBA {} Sukno%
}{%
Fernandez-Lopez%
\ \BBA {} Sukno%
}{%
{\protect \APACyear {2022}}%
}]{%
adriana2022alr}
\APACinsertmetastar {%
adriana2022alr}%
\begin{APACrefauthors}%
Fernandez-Lopez, A.%
\BCBT {}\ \BBA {} Sukno, F.M.%
\end{APACrefauthors}%
\unskip\
\newblock
\APACrefYearMonthDay{2022}{}{}.
\newblock
{\BBOQ}\APACrefatitle {End-to-{E}nd {L}ip-{R}eading {W}ithout {L}arge-{S}cale
  {D}ata} {End-to-{E}nd {L}ip-{R}eading {W}ithout {L}arge-{S}cale
  {D}ata}.{\BBCQ}
\newblock
\APACjournalVolNumPages{IEEE/ACM TASLP}{30}{}{2076-2090,}
\newblock
\begin{APACrefDOI} \doi{10.1109/TASLP.2022.3182274} \end{APACrefDOI}
\newblock

\newblock

\PrintBackRefs{\CurrentBib}

\bibitem [\protect \citeauthoryear {%
Fernández-López%
\ \BBA {} Sukno%
}{%
Fernández-López%
\ \BBA {} Sukno%
}{%
{\protect \APACyear {2017}}%
}]{%
fernandez2017optimizing}
\APACinsertmetastar {%
fernandez2017optimizing}%
\begin{APACrefauthors}%
Fernández-López, A.%
\BCBT {}\ \BBA {} Sukno, F.%
\end{APACrefauthors}%
\unskip\
\newblock
\APACrefYearMonthDay{2017}{}{}.
\newblock
{\BBOQ}\APACrefatitle {Optimizing Phoneme-to-Viseme Mapping for Continuous
  Lip-Reading in Spanish} {Optimizing phoneme-to-viseme mapping for continuous
  lip-reading in spanish}.{\BBCQ}
\newblock
 \APACrefbtitle {International Joint Conference on Computer Vision, Imaging and
  Computer Graphics} {International joint conference on computer vision,
  imaging and computer graphics}\ (\BPGS\ 305--328).
\newblock
\begin{APACrefURL} {\url{https://doi.org/10.1007/978-3-030-12209-6_15}}
  \end{APACrefURL}
\PrintBackRefs{\CurrentBib}

\bibitem [\protect \citeauthoryear {%
Gales%
\ \BBA {} Young%
}{%
Gales%
\ \BBA {} Young%
}{%
{\protect \APACyear {2008}}%
}]{%
gales2008application}
\APACinsertmetastar {%
gales2008application}%
\begin{APACrefauthors}%
Gales, M.%
\BCBT {}\ \BBA {} Young, S.%
\end{APACrefauthors}%
\unskip\
\newblock
\APACrefYear{2008}.
\newblock
\APACrefbtitle {The application of hidden Markov models in speech recognition}
  {The application of hidden markov models in speech recognition}.
\newblock
\APACaddressPublisher{}{Now Publishers Inc}.
\newblock
\begin{APACrefURL} {\url{http://doi.org/10.1561/2000000004}} \end{APACrefURL}
\PrintBackRefs{\CurrentBib}

\bibitem [\protect \citeauthoryear {%
Gimeno-G{\'o}mez%
\ \BBA {} Mart{\'\i}nez-Hinarejos%
}{%
Gimeno-G{\'o}mez%
\ \BBA {} Mart{\'\i}nez-Hinarejos%
}{%
{\protect \APACyear {2024}}%
}]{%
gimeno2024continuous}
\APACinsertmetastar {%
gimeno2024continuous}%
\begin{APACrefauthors}%
Gimeno-G{\'o}mez, D.%
\BCBT {}\ \BBA {} Mart{\'\i}nez-Hinarejos, C\BHBI D.%
\end{APACrefauthors}%
\unskip\
\newblock
\APACrefYearMonthDay{2024}{}{}.
\newblock
{\BBOQ}\APACrefatitle {{Continuous lipreading based on acoustic temporal
  alignments}} {{Continuous lipreading based on acoustic temporal
  alignments}}.{\BBCQ}
\newblock
\APACjournalVolNumPages{EURASIP Journal on Audio, Speech, and Music
  Processing}{2024}{1}{25,}
\newblock

\newblock

\PrintBackRefs{\CurrentBib}

\bibitem [\protect \citeauthoryear {%
Gimeno-Gómez%
\ \BBA {} Martínez-Hinarejos%
}{%
Gimeno-Gómez%
\ \BBA {} Martínez-Hinarejos%
}{%
{\protect \APACyear {2022}}%
}]{%
lrec2022liprtve}
\APACinsertmetastar {%
lrec2022liprtve}%
\begin{APACrefauthors}%
Gimeno-Gómez, D.%
\BCBT {}\ \BBA {} Martínez-Hinarejos, C\BHBI D.%
\end{APACrefauthors}%
\unskip\
\newblock
\APACrefYearMonthDay{2022}{}{}.
\newblock
{\BBOQ}\APACrefatitle {L{IP-RTVE}: {A}n {A}udiovisual {D}atabase for
  {C}ontinuous {S}panish in the {W}ild} {L{IP-RTVE}: {A}n {A}udiovisual
  {D}atabase for {C}ontinuous {S}panish in the {W}ild}.{\BBCQ}
\newblock
 \APACrefbtitle {{Proc. LREC}} {{Proc. LREC}}\ (\BPGS\ 2750--2758).
\newblock
\APACaddressPublisher{}{ELRA}.
\newblock
\begin{APACrefURL} {\url{https://aclanthology.org/2022.lrec-1.294}}
  \end{APACrefURL}
\PrintBackRefs{\CurrentBib}

\bibitem [\protect \citeauthoryear {%
Graves%
, Fern\'{a}ndez%
, Gomez%
\BCBL {}\ \BBA {} Schmidhuber%
}{%
Graves%
\ \protect \BOthers {.}}{%
{\protect \APACyear {2006}}%
}]{%
graves2006ctc}
\APACinsertmetastar {%
graves2006ctc}%
\begin{APACrefauthors}%
Graves, A.%
, Fern\'{a}ndez, S.%
, Gomez, F.%
\BCBL {} Schmidhuber, J.%
\end{APACrefauthors}%
\unskip\
\newblock
\APACrefYearMonthDay{2006}{}{}.
\newblock
{\BBOQ}\APACrefatitle {Connectionist Temporal Classification: Labelling
  Unsegmented Sequence Data with Recurrent Neural Networks} {Connectionist
  temporal classification: Labelling unsegmented sequence data with recurrent
  neural networks}.{\BBCQ}
\newblock
 \APACrefbtitle {23rd ICML} {23rd icml}\ (\BPG~369–376).
\newblock
\APACaddressPublisher{}{ACM}.
\newblock
\begin{APACrefURL} {\url{https://doi.org/10.1145/1143844.1143891}}
  \end{APACrefURL}
\PrintBackRefs{\CurrentBib}

\bibitem [\protect \citeauthoryear {%
Gulati%
\ \protect \BOthers {.}}{%
Gulati%
\ \protect \BOthers {.}}{%
{\protect \APACyear {2020}}%
}]{%
gulati20_interspeech}
\APACinsertmetastar {%
gulati20_interspeech}%
\begin{APACrefauthors}%
Gulati, A.%
, Qin, J.%
, Chiu, C.C.%
, Parmar, N.%
, Zhang, Y.%
, Yu, J.%
\BDBL {}Pang, R.%
\end{APACrefauthors}%
\unskip\
\newblock
\APACrefYearMonthDay{2020}{}{}.
\newblock
{\BBOQ}\APACrefatitle {{Conformer: Convolution-augmented Transformer for Speech
  Recognition}} {{Conformer: Convolution-augmented Transformer for Speech
  Recognition}}.{\BBCQ}
\newblock
 \APACrefbtitle {{Proc. Interspeech}} {{Proc. Interspeech}}\ (\BPGS\
  5036--5040).
\newblock
\begin{APACrefURL} {\url{https://doi.org/10.21437/Interspeech.2020-3015}}
  \end{APACrefURL}
\PrintBackRefs{\CurrentBib}

\bibitem [\protect \citeauthoryear {%
Haliassos%
, Zinonos%
, Mira%
, Petridis%
\BCBL {}\ \BBA {} Pantic%
}{%
Haliassos%
\ \protect \BOthers {.}}{%
{\protect \APACyear {2024}}%
}]{%
haliassos2024braven}
\APACinsertmetastar {%
haliassos2024braven}%
\begin{APACrefauthors}%
Haliassos, A.%
, Zinonos, A.%
, Mira, R.%
, Petridis, S.%
\BCBL {} Pantic, M.%
\end{APACrefauthors}%
\unskip\
\newblock
\APACrefYearMonthDay{2024}{}{}.
\newblock
{\BBOQ}\APACrefatitle {{BRAVEn: Improving Self-supervised pre-training for
  Visual and Auditory Speech Recognition}} {{BRAVEn: Improving Self-supervised
  pre-training for Visual and Auditory Speech Recognition}}.{\BBCQ}
\newblock
 \APACrefbtitle {ICASSP} {Icassp}\ (\BPG~11431-11435).
\PrintBackRefs{\CurrentBib}

\bibitem [\protect \citeauthoryear {%
Harte%
\ \BBA {} Gillen%
}{%
Harte%
\ \BBA {} Gillen%
}{%
{\protect \APACyear {2015}}%
}]{%
harte2015tcd}
\APACinsertmetastar {%
harte2015tcd}%
\begin{APACrefauthors}%
Harte, N.%
\BCBT {}\ \BBA {} Gillen, E.%
\end{APACrefauthors}%
\unskip\
\newblock
\APACrefYearMonthDay{2015}{}{}.
\newblock
{\BBOQ}\APACrefatitle {{TCD-TIMIT: An audio-visual corpus of continuous
  speech}} {{TCD-TIMIT: An audio-visual corpus of continuous speech}}.{\BBCQ}
\newblock
\APACjournalVolNumPages{IEEE Transactions on Multimedia}{17}{5}{603--615,}
\newblock
\begin{APACrefDOI} \doi{10.1109/TMM.2015.2407694} \end{APACrefDOI}
\newblock

\newblock

\PrintBackRefs{\CurrentBib}

\bibitem [\protect \citeauthoryear {%
He%
, Zhang%
, Ren%
\BCBL {}\ \BBA {} Sun%
}{%
He%
\ \protect \BOthers {.}}{%
{\protect \APACyear {2016}}%
}]{%
he2016resnet}
\APACinsertmetastar {%
he2016resnet}%
\begin{APACrefauthors}%
He, K.%
, Zhang, X.%
, Ren, S.%
\BCBL {} Sun, J.%
\end{APACrefauthors}%
\unskip\
\newblock
\APACrefYearMonthDay{2016}{}{}.
\newblock
{\BBOQ}\APACrefatitle {Deep Residual Learning for Image Recognition} {Deep
  residual learning for image recognition}.{\BBCQ}
\newblock
 \APACrefbtitle {{CVPR}} {{CVPR}}\ (\BPG~770-778).
\newblock
\begin{APACrefURL} {\url{https://doi.org/10.1109/CVPR.2016.90}}
  \end{APACrefURL}
\PrintBackRefs{\CurrentBib}

\bibitem [\protect \citeauthoryear {%
Higuchi%
, Inaguma%
, Watanabe%
, Ogawa%
\BCBL {}\ \BBA {} Kobayashi%
}{%
Higuchi%
\ \protect \BOthers {.}}{%
{\protect \APACyear {2021}}%
}]{%
higuchi2021maskctc}
\APACinsertmetastar {%
higuchi2021maskctc}%
\begin{APACrefauthors}%
Higuchi, Y.%
, Inaguma, H.%
, Watanabe, S.%
, Ogawa, T.%
\BCBL {} Kobayashi, T.%
\end{APACrefauthors}%
\unskip\
\newblock
\APACrefYearMonthDay{2021}{}{}.
\newblock
{\BBOQ}\APACrefatitle {{Improved Mask-CTC for Non-Autoregressive End-to-End
  ASR}} {{Improved Mask-CTC for Non-Autoregressive End-to-End ASR}}.{\BBCQ}
\newblock
 \APACrefbtitle {{ICASSP}} {{ICASSP}}\ (\BPG~8363-8367).
\PrintBackRefs{\CurrentBib}

\bibitem [\protect \citeauthoryear {%
Ivanko%
, Ryumin%
\BCBL {}\ \BBA {} Karpov%
}{%
Ivanko%
\ \protect \BOthers {.}}{%
{\protect \APACyear {2019}}%
}]{%
ivanko2019signlip}
\APACinsertmetastar {%
ivanko2019signlip}%
\begin{APACrefauthors}%
Ivanko, D.%
, Ryumin, D.%
\BCBL {} Karpov, A.%
\end{APACrefauthors}%
\unskip\
\newblock
\APACrefYearMonthDay{2019}{}{}.
\newblock
{\BBOQ}\APACrefatitle {{Automatic Lip-Reading of Hearing Impaired People}}
  {{Automatic Lip-Reading of Hearing Impaired People}}.{\BBCQ}
\newblock
\APACjournalVolNumPages{The International Archives of the Photogrammetry,
  Remote Sensing and Spatial Information Sciences}{XLII-2/W12}{}{97--101,}
\newblock
\begin{APACrefDOI} \doi{10.5194/isprs-archives-XLII-2-W12-97-2019}
  \end{APACrefDOI}
\newblock

\newblock

\PrintBackRefs{\CurrentBib}

\bibitem [\protect \citeauthoryear {%
Jha%
, Namboodiri%
\BCBL {}\ \BBA {} Jawahar%
}{%
Jha%
\ \protect \BOthers {.}}{%
{\protect \APACyear {2019}}%
}]{%
jha2019spotting}
\APACinsertmetastar {%
jha2019spotting}%
\begin{APACrefauthors}%
Jha, A.%
, Namboodiri, V.P.%
\BCBL {} Jawahar, C.V.%
\end{APACrefauthors}%
\unskip\
\newblock
\APACrefYearMonthDay{2019}{}{}.
\newblock
{\BBOQ}\APACrefatitle {{Spotting Words in Silent Speech Videos: A
  Retrieval-Based Approach}} {{Spotting Words in Silent Speech Videos: A
  Retrieval-Based Approach}}.{\BBCQ}
\newblock
\APACjournalVolNumPages{Machine Vision and Applications}{30}{}{217--229,}
\newblock

\newblock

\PrintBackRefs{\CurrentBib}

\bibitem [\protect \citeauthoryear {%
Kim%
, Yeo%
, Choi%
\BCBL {}\ \BBA {} Ro%
}{%
Kim%
\ \protect \BOthers {.}}{%
{\protect \APACyear {2023}}%
}]{%
kim2023lip}
\APACinsertmetastar {%
kim2023lip}%
\begin{APACrefauthors}%
Kim, M.%
, Yeo, J.H.%
, Choi, J.%
\BCBL {} Ro, Y.M.%
\end{APACrefauthors}%
\unskip\
\newblock
\APACrefYearMonthDay{2023}{}{}.
\newblock
{\BBOQ}\APACrefatitle {{Lip reading for low-resource languages by learning and
  combining general speech knowledge and language-specific knowledge}} {{Lip
  reading for low-resource languages by learning and combining general speech
  knowledge and language-specific knowledge}}.{\BBCQ}
\newblock
 \APACrefbtitle {ICCV} {Iccv}\ (\BPGS\ 15359--15371).
\PrintBackRefs{\CurrentBib}

\bibitem [\protect \citeauthoryear {%
Koller%
, Forster%
\BCBL {}\ \BBA {} Ney%
}{%
Koller%
\ \protect \BOthers {.}}{%
{\protect \APACyear {2015}}%
}]{%
koller2015continuous}
\APACinsertmetastar {%
koller2015continuous}%
\begin{APACrefauthors}%
Koller, O.%
, Forster, J.%
\BCBL {} Ney, H.%
\end{APACrefauthors}%
\unskip\
\newblock
\APACrefYearMonthDay{2015}{}{}.
\newblock
{\BBOQ}\APACrefatitle {{Continuous Sign Language Recognition: Towards Large
  Vocabulary Statistical Recognition Systems Handling Multiple Signers}}
  {{Continuous Sign Language Recognition: Towards Large Vocabulary Statistical
  Recognition Systems Handling Multiple Signers}}.{\BBCQ}
\newblock
\APACjournalVolNumPages{{Computer Vision and Image
  Understanding}}{141}{}{108--125,}
\newblock
\begin{APACrefDOI} \doi{10.1016/j.cviu.2015.09.013} \end{APACrefDOI}
\newblock

\newblock

\PrintBackRefs{\CurrentBib}

\bibitem [\protect \citeauthoryear {%
Laux%
\ \protect \BOthers {.}}{%
Laux%
\ \protect \BOthers {.}}{%
{\protect \APACyear {2023}}%
}]{%
laux2023care}
\APACinsertmetastar {%
laux2023care}%
\begin{APACrefauthors}%
Laux, H.%
, Hallawa, A.%
, Assis, J.C.S.%
, Schmeink, A.%
, Martin, L.%
\BCBL {} Peine, A.%
\end{APACrefauthors}%
\unskip\
\newblock
\APACrefYearMonthDay{2023}{}{}.
\newblock
{\BBOQ}\APACrefatitle {{Two-Stage Visual Speech Recognition for Intensive Care
  Patients}} {{Two-Stage Visual Speech Recognition for Intensive Care
  Patients}}.{\BBCQ}
\newblock
\APACjournalVolNumPages{Scientific Reports}{13}{1}{928,}
\newblock

\newblock

\PrintBackRefs{\CurrentBib}

\bibitem [\protect \citeauthoryear {%
Lee%
\ \BBA {} Watanabe%
}{%
Lee%
\ \BBA {} Watanabe%
}{%
{\protect \APACyear {2021}}%
}]{%
lee2021interctc}
\APACinsertmetastar {%
lee2021interctc}%
\begin{APACrefauthors}%
Lee, J.%
\BCBT {}\ \BBA {} Watanabe, S.%
\end{APACrefauthors}%
\unskip\
\newblock
\APACrefYearMonthDay{2021}{}{}.
\newblock
{\BBOQ}\APACrefatitle {{Intermediate Loss Regularization for CTC-Based Speech
  Recognition}} {{Intermediate Loss Regularization for CTC-Based Speech
  Recognition}}.{\BBCQ}
\newblock
 \APACrefbtitle {{ICASSP}} {{ICASSP}}\ (\BPG~6224-6228).
\PrintBackRefs{\CurrentBib}

\bibitem [\protect \citeauthoryear {%
Liao%
\ \protect \BOthers {.}}{%
Liao%
\ \protect \BOthers {.}}{%
{\protect \APACyear {2023}}%
}]{%
liao2023lightasd}
\APACinsertmetastar {%
liao2023lightasd}%
\begin{APACrefauthors}%
Liao, J.%
, Duan, H.%
, Feng, K.%
, Zhao, W.%
, Yang, Y.%
\BCBL {} Chen, L.%
\end{APACrefauthors}%
\unskip\
\newblock
\APACrefYearMonthDay{2023}{}{}.
\newblock
{\BBOQ}\APACrefatitle {{A Light Weight Model for Active Speaker Detection}} {{A
  Light Weight Model for Active Speaker Detection}}.{\BBCQ}
\newblock
 \APACrefbtitle {Proc. of the IEEE/CVF CVPR} {Proc. of the ieee/cvf cvpr}\
  (\BPG~22932-22941).
\PrintBackRefs{\CurrentBib}

\bibitem [\protect \citeauthoryear {%
Liu%
\ \protect \BOthers {.}}{%
Liu%
\ \protect \BOthers {.}}{%
{\protect \APACyear {2023}}%
}]{%
liu2023synthvsr}
\APACinsertmetastar {%
liu2023synthvsr}%
\begin{APACrefauthors}%
Liu, X.%
, Lakomkin, E.%
, Vougioukas, K.%
, Ma, P.%
, Chen, H.%
, Xie, R.%
\BDBL {}Fuegen, C.%
\end{APACrefauthors}%
\unskip\
\newblock
\APACrefYearMonthDay{2023}{}{}.
\newblock
{\BBOQ}\APACrefatitle {{SynthVSR: Scaling Up Visual Speech Recognition With
  Synthetic Supervision}} {{SynthVSR: Scaling Up Visual Speech Recognition With
  Synthetic Supervision}}.{\BBCQ}
\newblock
 \APACrefbtitle {CVPR} {Cvpr}\ (\BPGS\ 18806--18815).
\PrintBackRefs{\CurrentBib}

\bibitem [\protect \citeauthoryear {%
Loshchilov%
\ \BBA {} Hutter%
}{%
Loshchilov%
\ \BBA {} Hutter%
}{%
{\protect \APACyear {2019}}%
}]{%
loshchilov2017decoupled}
\APACinsertmetastar {%
loshchilov2017decoupled}%
\begin{APACrefauthors}%
Loshchilov, I.%
\BCBT {}\ \BBA {} Hutter, F.%
\end{APACrefauthors}%
\unskip\
\newblock
\APACrefYearMonthDay{2019}{}{}.
\newblock
{\BBOQ}\APACrefatitle {Decoupled Weight Decay Regularization} {Decoupled weight
  decay regularization}.{\BBCQ}
\newblock
 \APACrefbtitle {ICLR.} {Iclr.}
\newblock
\begin{APACrefURL} {\url{https://openreview.net/pdf?id=Bkg6RiCqY7}}
  \end{APACrefURL}
\PrintBackRefs{\CurrentBib}

\bibitem [\protect \citeauthoryear {%
Ma%
\ \protect \BOthers {.}}{%
Ma%
\ \protect \BOthers {.}}{%
{\protect \APACyear {2023}}%
}]{%
ma2023auto}
\APACinsertmetastar {%
ma2023auto}%
\begin{APACrefauthors}%
Ma, P.%
, Haliassos, A.%
, Fernandez-Lopez, A.%
, Chen, H.%
, Petridis, S.%
\BCBL {} Pantic, M.%
\end{APACrefauthors}%
\unskip\
\newblock
\APACrefYearMonthDay{2023}{}{}.
\newblock
{\BBOQ}\APACrefatitle {Auto-AVSR: Audio-Visual Speech Recognition with
  Automatic Labels} {Auto-avsr: Audio-visual speech recognition with automatic
  labels}.{\BBCQ}
\newblock
 \APACrefbtitle {ICASSP} {Icassp}\ (\BPG~1-5).
\PrintBackRefs{\CurrentBib}

\bibitem [\protect \citeauthoryear {%
Ma%
, Petridis%
\BCBL {}\ \BBA {} Pantic%
}{%
Ma%
\ \protect \BOthers {.}}{%
{\protect \APACyear {2021}}%
}]{%
maja2021conformers}
\APACinsertmetastar {%
maja2021conformers}%
\begin{APACrefauthors}%
Ma, P.%
, Petridis, S.%
\BCBL {} Pantic, M.%
\end{APACrefauthors}%
\unskip\
\newblock
\APACrefYearMonthDay{2021}{}{}.
\newblock
{\BBOQ}\APACrefatitle {End-To-End Audio-Visual Speech Recognition with
  Conformers} {End-to-end audio-visual speech recognition with
  conformers}.{\BBCQ}
\newblock
 \APACrefbtitle {I{CASSP}} {I{CASSP}}\ (\BPG~7613-7617).
\newblock
\begin{APACrefURL} {\url{https://doi.org/10.1109/ICASSP39728.2021.9414567}}
  \end{APACrefURL}
\PrintBackRefs{\CurrentBib}

\bibitem [\protect \citeauthoryear {%
Ma%
, Petridis%
\BCBL {}\ \BBA {} Pantic%
}{%
Ma%
\ \protect \BOthers {.}}{%
{\protect \APACyear {2022}}%
}]{%
ma2022visual}
\APACinsertmetastar {%
ma2022visual}%
\begin{APACrefauthors}%
Ma, P.%
, Petridis, S.%
\BCBL {} Pantic, M.%
\end{APACrefauthors}%
\unskip\
\newblock
\APACrefYearMonthDay{2022}{}{}.
\newblock
{\BBOQ}\APACrefatitle {Visual Speech Recognition for Multiple Languages in the
  Wild} {Visual speech recognition for multiple languages in the wild}.{\BBCQ}
\newblock
\APACjournalVolNumPages{Nature Machine Intelligence}{4}{11}{930--939,}
\newblock
\begin{APACrefDOI} \doi{10.1038/s42256-022-00550-z} \end{APACrefDOI}
\newblock

\newblock

\PrintBackRefs{\CurrentBib}

\bibitem [\protect \citeauthoryear {%
Manaris%
, Pellicoro%
, Pothering%
\BCBL {}\ \BBA {} Hodges%
}{%
Manaris%
\ \protect \BOthers {.}}{%
{\protect \APACyear {2006}}%
}]{%
manaris2006esperanto}
\APACinsertmetastar {%
manaris2006esperanto}%
\begin{APACrefauthors}%
Manaris, B.%
, Pellicoro, L.%
, Pothering, G.%
\BCBL {} Hodges, H.%
\end{APACrefauthors}%
\unskip\
\newblock
\APACrefYearMonthDay{2006}{}{}.
\newblock
{\BBOQ}\APACrefatitle {Investigating Esperanto's Statistical Proportions
  Relative to Other Languages Using Neural Networks and Zipf's Law}
  {Investigating esperanto's statistical proportions relative to other
  languages using neural networks and zipf's law}.{\BBCQ}
\newblock
 \APACrefbtitle {Proceedings of the 24th IASTED International Conference on
  Artificial Intelligence and Applications} {Proceedings of the 24th iasted
  international conference on artificial intelligence and applications}\
  (\BPG~102–108).
\newblock
\APACaddressPublisher{USA}{ACTA Press}.
\newblock
\begin{APACrefURL} {\url{https://doi.org/10.5555/1166890.1166908}}
  \end{APACrefURL}
\PrintBackRefs{\CurrentBib}

\bibitem [\protect \citeauthoryear {%
McGurk%
\ \BBA {} MacDonald%
}{%
McGurk%
\ \BBA {} MacDonald%
}{%
{\protect \APACyear {1976}}%
}]{%
mcgurk1976hearing}
\APACinsertmetastar {%
mcgurk1976hearing}%
\begin{APACrefauthors}%
McGurk, H.%
\BCBT {}\ \BBA {} MacDonald, J.%
\end{APACrefauthors}%
\unskip\
\newblock
\APACrefYearMonthDay{1976}{}{}.
\newblock
{\BBOQ}\APACrefatitle {Hearing lips and seeing voices} {Hearing lips and seeing
  voices}.{\BBCQ}
\newblock
\APACjournalVolNumPages{Nature}{264}{5588}{746--748,}
\newblock
\begin{APACrefDOI} \doi{10.1038/264746a0} \end{APACrefDOI}
\newblock

\newblock

\PrintBackRefs{\CurrentBib}

\bibitem [\protect \citeauthoryear {%
Musalia%
\ \protect \BOthers {.}}{%
Musalia%
\ \protect \BOthers {.}}{%
{\protect \APACyear {2023}}%
}]{%
musalia2023liopa}
\APACinsertmetastar {%
musalia2023liopa}%
\begin{APACrefauthors}%
Musalia, M.%
, Laha, S.%
, Cazalilla-Chica, J.%
, Allan, J.%
, Roach, L.%
, Twamley, J.%
\BDBL {}McAuley, D.F.%
\end{APACrefauthors}%
\unskip\
\newblock
\APACrefYearMonthDay{2023}{}{}.
\newblock
{\BBOQ}\APACrefatitle {A User Evaluation of Speech/Phrase Recognition Software
  in Critically Ill Patients: A DECIDE-AI Feasibility Study} {A user evaluation
  of speech/phrase recognition software in critically ill patients: A decide-ai
  feasibility study}.{\BBCQ}
\newblock
\APACjournalVolNumPages{{Critical Care}}{27}{1}{277,}
\newblock

\newblock

\PrintBackRefs{\CurrentBib}

\bibitem [\protect \citeauthoryear {%
Ott%
, Edunov%
, Grangier%
\BCBL {}\ \BBA {} Auli%
}{%
Ott%
\ \protect \BOthers {.}}{%
{\protect \APACyear {2018}}%
}]{%
ott2018accum}
\APACinsertmetastar {%
ott2018accum}%
\begin{APACrefauthors}%
Ott, M.%
, Edunov, S.%
, Grangier, D.%
\BCBL {} Auli, M.%
\end{APACrefauthors}%
\unskip\
\newblock
\APACrefYearMonthDay{2018}{}{}.
\newblock
{\BBOQ}\APACrefatitle {Scaling Neural Machine Translation} {Scaling neural
  machine translation}.{\BBCQ}
\newblock
 \APACrefbtitle {Proc. of the 3rd Conference on Machine Translation} {Proc. of
  the 3rd conference on machine translation}\ (\BPGS\ 1--9).
\newblock
\APACaddressPublisher{}{ACL}.
\newblock
\begin{APACrefURL} {\url{https://doi.org/10.18653/v1/W18-6301}}
  \end{APACrefURL}
\PrintBackRefs{\CurrentBib}

\bibitem [\protect \citeauthoryear {%
Park%
\ \protect \BOthers {.}}{%
Park%
\ \protect \BOthers {.}}{%
{\protect \APACyear {2024}}%
}]{%
park2024facetoface}
\APACinsertmetastar {%
park2024facetoface}%
\begin{APACrefauthors}%
Park, S.J.%
, Kim, C.W.%
, Rha, H.%
, Kim, M.%
, Hong, J.%
, Yeo, J.%
\BCBL {} Ro, Y.M.%
\end{APACrefauthors}%
\unskip\
\newblock
\APACrefYearMonthDay{2024}{}{}.
\newblock
{\BBOQ}\APACrefatitle {{Let's Go Real Talk: Spoken Dialogue Model for
  Face-to-Face Conversation}} {{Let's Go Real Talk: Spoken Dialogue Model for
  Face-to-Face Conversation}}.{\BBCQ}
\newblock
 \APACrefbtitle {Proc. of the 62nd ACL} {Proc. of the 62nd acl}\ (\BPGS\
  16334--16348).
\PrintBackRefs{\CurrentBib}

\bibitem [\protect \citeauthoryear {%
Piantadosi%
}{%
Piantadosi%
}{%
{\protect \APACyear {2014}}%
}]{%
piantadosi2014zipf}
\APACinsertmetastar {%
piantadosi2014zipf}%
\begin{APACrefauthors}%
Piantadosi, S.T.%
\end{APACrefauthors}%
\unskip\
\newblock
\APACrefYearMonthDay{2014}{}{}.
\newblock
{\BBOQ}\APACrefatitle {Zipf’s word frequency law in natural language: A
  critical review and future directions} {Zipf’s word frequency law in
  natural language: A critical review and future directions}.{\BBCQ}
\newblock
\APACjournalVolNumPages{Psychonomic bulletin \& review}{21}{}{1112--1130,}
\newblock
\begin{APACrefDOI} \doi{10.3758/s13423-014-0585-6} \end{APACrefDOI}
\newblock

\newblock

\PrintBackRefs{\CurrentBib}

\bibitem [\protect \citeauthoryear {%
Potamianos%
, Neti%
, Gravier%
, Garg%
\BCBL {}\ \BBA {} Senior%
}{%
Potamianos%
\ \protect \BOthers {.}}{%
{\protect \APACyear {2003}}%
}]{%
potamianos2003recent}
\APACinsertmetastar {%
potamianos2003recent}%
\begin{APACrefauthors}%
Potamianos, G.%
, Neti, C.%
, Gravier, G.%
, Garg, A.%
\BCBL {} Senior, A.%
\end{APACrefauthors}%
\unskip\
\newblock
\APACrefYearMonthDay{2003}{}{}.
\newblock
{\BBOQ}\APACrefatitle {Recent advances in the automatic recognition of
  audiovisual speech} {Recent advances in the automatic recognition of
  audiovisual speech}.{\BBCQ}
\newblock
\APACjournalVolNumPages{Proc. of the IEEE}{91}{9}{1306--1326,}
\newblock
\begin{APACrefDOI} \doi{10.1109/JPROC.2003.817150} \end{APACrefDOI}
\newblock

\newblock

\PrintBackRefs{\CurrentBib}

\bibitem [\protect \citeauthoryear {%
K.~Prajwal%
, Mukhopadhyay%
, Namboodiri%
\BCBL {}\ \BBA {} Jawahar%
}{%
K.~Prajwal%
\ \protect \BOthers {.}}{%
{\protect \APACyear {2020}}%
}]{%
prajwal2020lip}
\APACinsertmetastar {%
prajwal2020lip}%
\begin{APACrefauthors}%
Prajwal, K.%
, Mukhopadhyay, R.%
, Namboodiri, V.%
\BCBL {} Jawahar, C.V.%
\end{APACrefauthors}%
\unskip\
\newblock
\APACrefYearMonthDay{2020}{}{}.
\newblock
{\BBOQ}\APACrefatitle {A Lip Sync Expert Is All You Need for Speech to Lip
  Generation in the Wild} {A lip sync expert is all you need for speech to lip
  generation in the wild}.{\BBCQ}
\newblock
 \APACrefbtitle {Proceedings of the 28th ACM international conference on
  multimedia} {Proceedings of the 28th acm international conference on
  multimedia}\ (\BPGS\ 484--492).
\PrintBackRefs{\CurrentBib}

\bibitem [\protect \citeauthoryear {%
K.R.~Prajwal%
, Afouras%
\BCBL {}\ \BBA {} Zisserman%
}{%
K.R.~Prajwal%
\ \protect \BOthers {.}}{%
{\protect \APACyear {2022}}%
}]{%
prajwal2021sub}
\APACinsertmetastar {%
prajwal2021sub}%
\begin{APACrefauthors}%
Prajwal, K.R.%
, Afouras, T.%
\BCBL {} Zisserman, A.%
\end{APACrefauthors}%
\unskip\
\newblock
\APACrefYearMonthDay{2022}{}{}.
\newblock
{\BBOQ}\APACrefatitle {Sub-Word Level Lip Reading With Visual Attention}
  {Sub-word level lip reading with visual attention}.{\BBCQ}
\newblock
 \APACrefbtitle {{CVPR}} {{CVPR}}\ (\BPG~5162-5172).
\newblock
\begin{APACrefURL}
  {\url{https://openaccess.thecvf.com/content/CVPR2022/html/Prajwal_Sub-Word_Level_Lip_Reading_With_Visual_Attention_CVPR_2022_paper.html}}
  \end{APACrefURL}
\PrintBackRefs{\CurrentBib}

\bibitem [\protect \citeauthoryear {%
Pratap%
, Xu%
, Sriram%
, Synnaeve%
\BCBL {}\ \BBA {} Collobert%
}{%
Pratap%
\ \protect \BOthers {.}}{%
{\protect \APACyear {2020}}%
}]{%
pratap20interspeech}
\APACinsertmetastar {%
pratap20interspeech}%
\begin{APACrefauthors}%
Pratap, V.%
, Xu, Q.%
, Sriram, A.%
, Synnaeve, G.%
\BCBL {} Collobert, R.%
\end{APACrefauthors}%
\unskip\
\newblock
\APACrefYearMonthDay{2020}{}{}.
\newblock
{\BBOQ}\APACrefatitle {{MLS: A Large-Scale Multilingual Dataset for Speech
  Research}} {{MLS: A Large-Scale Multilingual Dataset for Speech
  Research}}.{\BBCQ}
\newblock
 \APACrefbtitle {Proc. Interspeech} {Proc. interspeech}\ (\BPGS\ 2757--2761).
\newblock
\begin{APACrefURL} {\url{https://doi.org/10.21437/Interspeech.2020-2826}}
  \end{APACrefURL}
\PrintBackRefs{\CurrentBib}

\bibitem [\protect \citeauthoryear {%
Ramachandran%
, Zoph%
\BCBL {}\ \BBA {} Le%
}{%
Ramachandran%
\ \protect \BOthers {.}}{%
{\protect \APACyear {2017}}%
}]{%
swish2017prajit}
\APACinsertmetastar {%
swish2017prajit}%
\begin{APACrefauthors}%
Ramachandran, P.%
, Zoph, B.%
\BCBL {} Le, Q.V.%
\end{APACrefauthors}%
\unskip\
\newblock
\APACrefYearMonthDay{2017}{}{}.
\newblock
{\BBOQ}\APACrefatitle {Searching for Activation Functions} {Searching for
  activation functions}.{\BBCQ}
\newblock
\APACjournalVolNumPages{arXiv preprint arXiv:1710.05941}{}{}{,}
\newblock
\begin{APACrefURL} {\url{https://arxiv.org/abs/1710.05941}} \end{APACrefURL}
\newblock

\newblock

\PrintBackRefs{\CurrentBib}

\bibitem [\protect \citeauthoryear {%
Salesky%
\ \protect \BOthers {.}}{%
Salesky%
\ \protect \BOthers {.}}{%
{\protect \APACyear {2021}}%
}]{%
salesky21_interspeech}
\APACinsertmetastar {%
salesky21_interspeech}%
\begin{APACrefauthors}%
Salesky, E.%
, Wiesner, M.%
, Bremerman, J.%
, Cattoni, R.%
, Negri, M.%
, Turchi, M.%
\BDBL {}Post, M.%
\end{APACrefauthors}%
\unskip\
\newblock
\APACrefYearMonthDay{2021}{}{}.
\newblock
{\BBOQ}\APACrefatitle {{The Multilingual TEDx Corpus for Speech Recognition and
  Translation}} {{The Multilingual TEDx Corpus for Speech Recognition and
  Translation}}.{\BBCQ}
\newblock
 \APACrefbtitle {Proc. Interspeech} {Proc. interspeech}\ (\BPGS\ 3655--3659).
\newblock
\begin{APACrefURL} {\url{https//doi.org/10.21437/Interspeech.2021-11}}
  \end{APACrefURL}
\PrintBackRefs{\CurrentBib}

\bibitem [\protect \citeauthoryear {%
Shi%
, Hsu%
, Lakhotia%
\BCBL {}\ \BBA {} Mohamed%
}{%
Shi%
\ \protect \BOthers {.}}{%
{\protect \APACyear {2022}}%
}]{%
shi2022learning}
\APACinsertmetastar {%
shi2022learning}%
\begin{APACrefauthors}%
Shi, B.%
, Hsu, W.N.%
, Lakhotia, K.%
\BCBL {} Mohamed, A.%
\end{APACrefauthors}%
\unskip\
\newblock
\APACrefYearMonthDay{2022}{}{}.
\newblock
{\BBOQ}\APACrefatitle {Learning audio-visual speech representation by masked
  multimodal cluster prediction} {Learning audio-visual speech representation
  by masked multimodal cluster prediction}.{\BBCQ}
\newblock
\APACjournalVolNumPages{arXiv preprint arXiv:2201.02184}{}{}{,}
\newblock
\begin{APACrefURL} {\url{https://doi.org/10.48550/arXiv.2201.02184}}
  \end{APACrefURL}
\newblock

\newblock

\PrintBackRefs{\CurrentBib}

\bibitem [\protect \citeauthoryear {%
Smith%
\ \BBA {} Topin%
}{%
Smith%
\ \BBA {} Topin%
}{%
{\protect \APACyear {2019}}%
}]{%
leslie2019onecycle}
\APACinsertmetastar {%
leslie2019onecycle}%
\begin{APACrefauthors}%
Smith, L.N.%
\BCBT {}\ \BBA {} Topin, N.%
\end{APACrefauthors}%
\unskip\
\newblock
\APACrefYearMonthDay{2019}{}{}.
\newblock
{\BBOQ}\APACrefatitle {{Super-convergence: very fast training of neural
  networks using large learning rates}} {{Super-convergence: very fast training
  of neural networks using large learning rates}}.{\BBCQ}
\newblock
 \APACrefbtitle {AI and ML for Multi-Domain Operations Applications} {Ai and ml
  for multi-domain operations applications}\ (\BVOL\ 11006, \BPGS\ 369--386).
\newblock
\begin{APACrefURL} {\url{https://doi.org/10.1117/12.2520589}} \end{APACrefURL}
\PrintBackRefs{\CurrentBib}

\bibitem [\protect \citeauthoryear {%
Son~Chung%
, Senior%
, Vinyals%
\BCBL {}\ \BBA {} Zisserman%
}{%
Son~Chung%
\ \protect \BOthers {.}}{%
{\protect \APACyear {2017}}%
}]{%
son2017lrs2}
\APACinsertmetastar {%
son2017lrs2}%
\begin{APACrefauthors}%
Son~Chung, J.%
, Senior, A.%
, Vinyals, O.%
\BCBL {} Zisserman, A.%
\end{APACrefauthors}%
\unskip\
\newblock
\APACrefYearMonthDay{2017}{}{}.
\newblock
{\BBOQ}\APACrefatitle {Lip reading sentences in the wild} {Lip reading
  sentences in the wild}.{\BBCQ}
\newblock
 \APACrefbtitle {{CVPR}} {{CVPR}}\ (\BPGS\ 6447--6456).
\newblock
\begin{APACrefURL}
  {\url{https://openaccess.thecvf.com/content_cvpr_2017/html/Chung_Lip_Reading_Sentences_CVPR_2017_paper.html}}
  \end{APACrefURL}
\PrintBackRefs{\CurrentBib}

\bibitem [\protect \citeauthoryear {%
Stafylakis%
\ \BBA {} Tzimiropoulos%
}{%
Stafylakis%
\ \BBA {} Tzimiropoulos%
}{%
{\protect \APACyear {2018}}%
}]{%
stafylakis2018zero}
\APACinsertmetastar {%
stafylakis2018zero}%
\begin{APACrefauthors}%
Stafylakis, T.%
\BCBT {}\ \BBA {} Tzimiropoulos, G.%
\end{APACrefauthors}%
\unskip\
\newblock
\APACrefYearMonthDay{2018}{}{}.
\newblock
{\BBOQ}\APACrefatitle {{Zero-Shot Keyword Spotting for Visual Speech
  Recognition In-the-Wild}} {{Zero-Shot Keyword Spotting for Visual Speech
  Recognition In-the-Wild}}.{\BBCQ}
\newblock
 \APACrefbtitle {Proc. of ECCV} {Proc. of eccv}\ (\BPGS\ 513--529).
\PrintBackRefs{\CurrentBib}

\bibitem [\protect \citeauthoryear {%
Tao%
\ \protect \BOthers {.}}{%
Tao%
\ \protect \BOthers {.}}{%
{\protect \APACyear {2021}}%
}]{%
tao2021talknetasd}
\APACinsertmetastar {%
tao2021talknetasd}%
\begin{APACrefauthors}%
Tao, R.%
, Pan, Z.%
, Das, R.%
, Qian, X.%
, Shou, M.%
\BCBL {} Li, h.%
\end{APACrefauthors}%
\unskip\
\newblock
\APACrefYearMonthDay{2021}{}{}.
\newblock
{\BBOQ}\APACrefatitle {{Is Someone Speaking? Exploring Long-Term Temporal
  Features for Audio-Visual Active Speaker Detection}} {{Is Someone Speaking?
  Exploring Long-Term Temporal Features for Audio-Visual Active Speaker
  Detection}}.{\BBCQ}
\newblock
 \APACrefbtitle {Proc. of the 29th ACM International Conference on Multimedia}
  {Proc. of the 29th acm international conference on multimedia}\
  (\BPG~3927–3935).
\newblock
\APACaddressPublisher{}{Association for Computing Machinery}.
\PrintBackRefs{\CurrentBib}

\bibitem [\protect \citeauthoryear {%
Thangthai%
}{%
Thangthai%
}{%
{\protect \APACyear {2018}}%
}]{%
thangthai2018computer}
\APACinsertmetastar {%
thangthai2018computer}%
\begin{APACrefauthors}%
Thangthai, K.%
\end{APACrefauthors}%
\unskip\
\newblock
\APACrefYear{2018}.
\unskip\
\newblock
\APACrefbtitle {Computer lipreading via hybrid deep neural network hidden
  Markov models} {Computer lipreading via hybrid deep neural network hidden
  markov models}\ \APACtypeAddressSchool {\BPhD}{}{University of East Anglia}.
\unskip\
\newblock
\begin{APACrefURL} {\url{https://ueaeprints.uea.ac.uk/id/eprint/69215}}
  \end{APACrefURL}
\PrintBackRefs{\CurrentBib}

\bibitem [\protect \citeauthoryear {%
Theobald%
, Harvey%
, Cox%
, Lewis%
\BCBL {}\ \BBA {} Owen%
}{%
Theobald%
\ \protect \BOthers {.}}{%
{\protect \APACyear {2006}}%
}]{%
theobald2006law}
\APACinsertmetastar {%
theobald2006law}%
\begin{APACrefauthors}%
Theobald, B.J.%
, Harvey, R.%
, Cox, S.J.%
, Lewis, C.%
\BCBL {} Owen, G.P.%
\end{APACrefauthors}%
\unskip\
\newblock
\APACrefYearMonthDay{2006}{}{}.
\newblock
{\BBOQ}\APACrefatitle {{Lip-Reading Enhancement for Law Enforcement}}
  {{Lip-Reading Enhancement for Law Enforcement}}.{\BBCQ}
\newblock
 \APACrefbtitle {Optics and Photonics for Counterterrorism and Crime Fighting
  II} {Optics and photonics for counterterrorism and crime fighting ii}\
  (\BVOL\ 6402, \BPGS\ 24--32).
\PrintBackRefs{\CurrentBib}

\bibitem [\protect \citeauthoryear {%
Vaswani%
\ \protect \BOthers {.}}{%
Vaswani%
\ \protect \BOthers {.}}{%
{\protect \APACyear {2017}}%
}]{%
vaswani2017attention}
\APACinsertmetastar {%
vaswani2017attention}%
\begin{APACrefauthors}%
Vaswani, A.%
, Shazeer, N.%
, Parmar, N.%
, Uszkoreit, J.%
, Jones, L.%
, Gomez, A.N.%
\BDBL {}Polosukhin, I.%
\end{APACrefauthors}%
\unskip\
\newblock
\APACrefYearMonthDay{2017}{}{}.
\newblock
{\BBOQ}\APACrefatitle {Attention is all you need} {Attention is all you
  need}.{\BBCQ}
\newblock
\APACjournalVolNumPages{NeurIPS}{30}{}{6000--6010,}
\newblock
\begin{APACrefURL} {\url{https://dl.acm.org/doi/10.5555/3295222.3295349}}
  \end{APACrefURL}
\newblock

\newblock

\PrintBackRefs{\CurrentBib}

\bibitem [\protect \citeauthoryear {%
Watanabe%
\ \protect \BOthers {.}}{%
Watanabe%
\ \protect \BOthers {.}}{%
{\protect \APACyear {2018}}%
}]{%
watanabe18_interspeech}
\APACinsertmetastar {%
watanabe18_interspeech}%
\begin{APACrefauthors}%
Watanabe, S.%
, Hori, T.%
, Karita, S.%
, Hayashi, T.%
, Nishitoba, J.%
, Unno, Y.%
\BDBL {}Ochiai, T.%
\end{APACrefauthors}%
\unskip\
\newblock
\APACrefYearMonthDay{2018}{}{}.
\newblock
{\BBOQ}\APACrefatitle {{ESPnet: End-to-End Speech Processing Toolkit}}
  {{ESPnet: End-to-End Speech Processing Toolkit}}.{\BBCQ}
\newblock
 \APACrefbtitle {Proc. Interspeech} {Proc. interspeech}\ (\BPGS\ 2207--2211).
\newblock
\begin{APACrefURL} {\url{https://doi.org/10.21437/Interspeech.2018-1456}}
  \end{APACrefURL}
\PrintBackRefs{\CurrentBib}

\bibitem [\protect \citeauthoryear {%
Watanabe%
, Hori%
, Kim%
, Hershey%
\BCBL {}\ \BBA {} Hayashi%
}{%
Watanabe%
\ \protect \BOthers {.}}{%
{\protect \APACyear {2017}}%
}]{%
watanabe2017ctcattention}
\APACinsertmetastar {%
watanabe2017ctcattention}%
\begin{APACrefauthors}%
Watanabe, S.%
, Hori, T.%
, Kim, S.%
, Hershey, J.R.%
\BCBL {} Hayashi, T.%
\end{APACrefauthors}%
\unskip\
\newblock
\APACrefYearMonthDay{2017}{}{}.
\newblock
{\BBOQ}\APACrefatitle {Hybrid CTC/Attention Architecture for End-to-End Speech
  Recognition} {Hybrid ctc/attention architecture for end-to-end speech
  recognition}.{\BBCQ}
\newblock
\APACjournalVolNumPages{IEEE JSTSP}{11}{8}{1240-1253,}
\newblock
\begin{APACrefDOI} \doi{10.1109/JSTSP.2017.2763455} \end{APACrefDOI}
\newblock

\newblock

\PrintBackRefs{\CurrentBib}

\bibitem [\protect \citeauthoryear {%
Wei%
\ \protect \BOthers {.}}{%
Wei%
\ \protect \BOthers {.}}{%
{\protect \APACyear {2023}}%
}]{%
wei2023sim}
\APACinsertmetastar {%
wei2023sim}%
\begin{APACrefauthors}%
Wei, G.%
, Duan, Z.%
, Li, S.%
, Yang, G.%
, Yu, X.%
\BCBL {} Li, J.%
\end{APACrefauthors}%
\unskip\
\newblock
\APACrefYearMonthDay{2023}{}{}.
\newblock
{\BBOQ}\APACrefatitle {{Sim-T: Simplify the Transformer Network by Multiplexing
  Technique for Speech Recognition}} {{Sim-T: Simplify the Transformer Network
  by Multiplexing Technique for Speech Recognition}}.{\BBCQ}
\newblock
\APACjournalVolNumPages{{arXiv preprint arXiv:2304.04991}}{}{}{,}
\newblock

\newblock

\PrintBackRefs{\CurrentBib}

\bibitem [\protect \citeauthoryear {%
Yang%
\ \protect \BOthers {.}}{%
Yang%
\ \protect \BOthers {.}}{%
{\protect \APACyear {2019}}%
}]{%
lrw2019chinese}
\APACinsertmetastar {%
lrw2019chinese}%
\begin{APACrefauthors}%
Yang, S.%
, Zhang, Y.%
, Feng, D.%
, Yang, M.%
, Wang, C.%
, Xiao, J.%
\BDBL {}Chen, X.%
\end{APACrefauthors}%
\unskip\
\newblock
\APACrefYearMonthDay{2019}{}{}.
\newblock
{\BBOQ}\APACrefatitle {{LRW-1000: A Naturally-Distributed Large-Scale Benchmark
  for Lip Reading in the Wild}} {{LRW-1000: A Naturally-Distributed Large-Scale
  Benchmark for Lip Reading in the Wild}}.{\BBCQ}
\newblock
 \APACrefbtitle {14th IEEE International Conference on Automatic Face \&
  Gesture Recognition} {14th ieee international conference on automatic face \&
  gesture recognition}\ (\BPG~1-8).
\PrintBackRefs{\CurrentBib}

\bibitem [\protect \citeauthoryear {%
Yeo%
, Kim%
, Watanabe%
\BCBL {}\ \BBA {} Ro%
}{%
Yeo%
\ \protect \BOthers {.}}{%
{\protect \APACyear {2024}}%
}]{%
yeo2024limited}
\APACinsertmetastar {%
yeo2024limited}%
\begin{APACrefauthors}%
Yeo, J.H.%
, Kim, M.%
, Watanabe, S.%
\BCBL {} Ro, Y.M.%
\end{APACrefauthors}%
\unskip\
\newblock
\APACrefYearMonthDay{2024}{}{}.
\newblock
{\BBOQ}\APACrefatitle {Visual Speech Recognition for Languages with Limited
  Labeled Data Using Automatic Labels from Whisper} {Visual speech recognition
  for languages with limited labeled data using automatic labels from
  whisper}.{\BBCQ}
\newblock
 \APACrefbtitle {ICASSP} {Icassp}\ (\BPG~10471-10475).
\PrintBackRefs{\CurrentBib}

\bibitem [\protect \citeauthoryear {%
Zadeh%
\ \protect \BOthers {.}}{%
Zadeh%
\ \protect \BOthers {.}}{%
{\protect \APACyear {2020}}%
}]{%
zadeh2020moseas}
\APACinsertmetastar {%
zadeh2020moseas}%
\begin{APACrefauthors}%
Zadeh, A.B.%
, Cao, Y.%
, Hessner, S.%
, Liang, P.P.%
, Poria, S.%
\BCBL {} Morency, L\BHBI P.%
\end{APACrefauthors}%
\unskip\
\newblock
\APACrefYearMonthDay{2020}{}{}.
\newblock
{\BBOQ}\APACrefatitle {C{MU-MOSEAS}: A Multimodal Language Dataset for Spanish,
  Portuguese, German and French} {C{MU-MOSEAS}: A multimodal language dataset
  for spanish, portuguese, german and french}.{\BBCQ}
\newblock
 \APACrefbtitle {{EMNLP}} {{EMNLP}}\ (\BPGS\ 1801--1812).
\newblock
\begin{APACrefURL} {\url{https://doi.org/10.18653/v1/2020.emnlp-main.141}}
  \end{APACrefURL}
\PrintBackRefs{\CurrentBib}

\bibitem [\protect \citeauthoryear {%
Zhang%
, Yang%
, Xiao%
, Shan%
\BCBL {}\ \BBA {} Chen%
}{%
Zhang%
\ \protect \BOthers {.}}{%
{\protect \APACyear {2020}}%
}]{%
zhang2020rois}
\APACinsertmetastar {%
zhang2020rois}%
\begin{APACrefauthors}%
Zhang, Y.%
, Yang, S.%
, Xiao, J.%
, Shan, S.%
\BCBL {} Chen, X.%
\end{APACrefauthors}%
\unskip\
\newblock
\APACrefYearMonthDay{2020}{}{}.
\newblock
{\BBOQ}\APACrefatitle {Can We Read Speech Beyond the Lips? Rethinking RoI
  Selection for Deep Visual Speech Recognition} {Can we read speech beyond the
  lips? rethinking roi selection for deep visual speech recognition}.{\BBCQ}
\newblock
 \APACrefbtitle {15th IEEE FG} {15th ieee fg}\ (\BPG~356-363).
\PrintBackRefs{\CurrentBib}

\bibitem [\protect \citeauthoryear {%
Zipf%
}{%
Zipf%
}{%
{\protect \APACyear {1936}}%
}]{%
zipf1936law}
\APACinsertmetastar {%
zipf1936law}%
\begin{APACrefauthors}%
Zipf, G.K.%
\end{APACrefauthors}%
\unskip\
\newblock
\APACrefYearMonthDay{1936}{}{}.
\newblock
\APACrefbtitle {The Psychobiology of Language.} {The psychobiology of
  language.}
\newblock
\APACaddressPublisher{}{Houghton, Mifflin}.
\newblock
\begin{APACrefURL} {\url{https://psycnet.apa.org/record/1935-04756-000}}
  \end{APACrefURL}
\PrintBackRefs{\CurrentBib}

\bibitem [\protect \citeauthoryear {%
Zipf%
}{%
Zipf%
}{%
{\protect \APACyear {1949}}%
}]{%
zipf1949human}
\APACinsertmetastar {%
zipf1949human}%
\begin{APACrefauthors}%
Zipf, G.K.%
\end{APACrefauthors}%
\unskip\
\newblock
\APACrefYearMonthDay{1949}{}{}.
\newblock
\APACrefbtitle {Human Behavior and the Principle of Least Effort.} {Human
  behavior and the principle of least effort.}
\newblock
\APACaddressPublisher{}{Addison-Wesley Press}.
\newblock
\begin{APACrefURL} {\url{https://psycnet.apa.org/record/1950-00412-000}}
  \end{APACrefURL}
\PrintBackRefs{\CurrentBib}

\end{thebibliography}

\end{document}